\documentclass[accepted]{template}
\usepackage{natbib}
\usepackage{mathtools}
\usepackage{booktabs}
\usepackage{tikz}
\usepackage{amsmath}
\usepackage{caption}
\usepackage{subcaption}
\usepackage{amsfonts}
\DeclareMathOperator*{\argmax}{argmax}

\providecommand{\keywords}[1]{\textbf{\textit{Keywords---}} #1}

\title{Assessing the Impact of Context Inference Error and Partial Observability\\
on RL Methods for Just-In-Time Adaptive Interventions}

\author[1]{\href{mailto:<karine@cs.umass.edu>}{Karine Karine}}
\author[2]{\href{mailto:<klasnja@umich.edu>}{Predrag Klasnja}}
\author[3]{\href{mailto:<samurphy@g.harvard.edu>}{Susan A. Murphy}}
\author[1]{\href{mailto:<marlin@cs.umass.edu>}{Benjamin M. Marlin}}
\affil[1]{University of Massachusetts Amherst}
\affil[2]{University of Michigan}
\affil[3]{Harvard University}
 
\begin{document}

\maketitle

\begin{abstract}
Just-in-Time Adaptive Interventions (JITAIs) are a class of personalized health interventions developed within the behavioral science community. JITAIs aim to provide the right type and amount of support by iteratively selecting a sequence of intervention options from a pre-defined set of components in response to each individual's time varying state. In this work, we explore the application of reinforcement learning methods to the problem of learning intervention option selection policies. We study the effect of context inference error and partial observability on the ability to learn effective policies. Our results show that the propagation of uncertainty from context inferences is critical to improving intervention efficacy as context uncertainty increases, while policy gradient algorithms can provide remarkable robustness to partially observed behavioral state information.
\end{abstract}

\begin{footnotesize}
\end{footnotesize}

\vspace{-1.5mm}
\section{Introduction}
\label{sec:intro}

Just-in-Time Adaptive Interventions (or JITAIs) are a class of personalized health intervention developed within the behavioral science community \citep{nahum2018just,hardeman2019systematic, battalio2021sense2stop, yang2023just, perski2022technology}. The primary goal of JITAIs is to provide the right type and amount of support for each individual as their personal and environmental context varies over time \citep{nahum2018just}. JITAIs aim to accomplish this goal by using decision rules to select from among a collection of possible intervention options based on observed and inferred dimensions of an individual's state. 

While current JITAI's and related adaptive intervention designs leverage increasingly sophisticated wearable sensors and machine-learning based context inference methods \citep{battalio2021sense2stop}, JITAI decision rules are still largely developed using an expert systems approach \citep{perski2022technology}. In this work, we investigate the application of neural network-based reinforcement learning (RL) methods \citep{Williams-92, Mnih-13}  to the problem of learning intervention option selection policies for JITAIs using a novel simulation environment that captures key behavioral concepts including habituation and risk of disengagement with an intervention.

We focus on two foundational issues with the application of RL algorithms to JITAIs. First, we investigate the impact of context inference error on the performance of learned policies. Second, we investigate the impact of non-observability of psychological state variables on policy learning. We note that neither of these issues has received attention in prior work and current JITAIs routinely leverage machine learning-based context inferences that discard prediction uncertainty. 

Our primary contributions are: (1) the development of a physical activity JITAI simulation environment that captures key aspects of the dynamics of behavior in the context of adaptive interventions; and (2) the quantitative evaluation of the impact of context inference error, context inference uncertainty and partial observability on the performance of policies learned using different categories of reinforcement learning approaches including policy gradient methods and value function methods. 

Our results show that policies that leverage context inference probabilities as features can significantly outperform policies that use only the most likely context value. Second, our results show that non-observability of psychological state variables has a drastic impact on the quality of polices learned using value function methods, but a significantly more modest effect on policy gradient methods. These results have important implications for the design of RL methods for use in JITAI applications.

The remainder of this paper is organized as follows. In Section \ref{sec:related_work} we provide background on JITAIs and reinforcement learning methods. In Section \ref{sec:methods} we present the methods used in our experiments including the description of the physical activity JITAI simulation environment. In Section \ref{sec:experiments} we present experiments and results. We conclude with a discussion in Section \ref{sec:conclusions}.

\section{Background and Related Work}\label{sec:related_work}

In this section we provide a brief overview of research on JITAIs and background on reinforcement learning methods. 

\subsection{Just-in-Time Adaptive Interventions}

As noted in the introduction, JITAIs are a class of personalized health intervention developed within the behavioral science community that aim to provide the right type and amount of support for each individual as their personal and environmental context varies over time \citep{nahum2018just}. JITAI's and related adaptive study design have been applied in multiple critical health domains including physical activity \citep{hardeman2019systematic}, smoking cessation \citep{battalio2021sense2stop, yang2023just} and addiction \citep{perski2022technology}. 

JITAIs are comprised of three main parts: the set of intervention components that can be provided to an individual and the specific intervention options within each component; a set of decision time points that determine when intervention components can be provided to an individual, and a policy that determines which intervention option to select for a given individual in a given context. Many current JITAIs are sophisticated cloud-supported mobile software applications that leverage a variety of intervention components from planning to goal setting to contextually tailored messaging and content delivered from auxiliary apps (such as mindfulness and stress reduction exercises) \citep{perski2022technology, spruijt2022advancing}. 

While early JITAIs were largely based on self-report of context information, current JITAIs are increasingly making use of machine learning-based context inferences derived from data collected from smart phones and wearable sensors. For example, recent work in adaptive intervention design for smoking cessation support \citep{battalio2021sense2stop} leverages customized wearables \citep{ertin2011autosense,kwon2021validity} and machine learning models for the detection of stress \citep{hovsepian2015cstress} as well as smoking lapse \citep{saleheen2015puffmarker}. 

Despite the sophistication of JITAIs as software applications, the complexity of component and option selection policies has remained relatively limited. While the policies are adaptive in the sense of selecting different content in different contexts, the context-to-content mappings are often hand-designed by the intervention designers. While this allows intervention designers to build selection policies that are based on behavioral theory, there is significant need for methods that can refine expert policies as well as learn de novo policies from data. 

To this end, a number of domains where JITAIs are being deployed admit meaningful and continuously measurable proximal outcomes that can be used as a reward signal for reinforcement learning algorithms. For example, in the physical activity domain, wearable activity tracking devices such as FitBit devices and smart watches can be used to detect both the duration of sedentary episodes as well as steps \citep{spruijt2022advancing}. We turn next to a brief review of reinforcement learning and return to a discussion of the challenges of applying RL methods in the JITAI context at the end of this section.

\subsection{Reinforcement Learning}

The goal of reinforcement learning (RL) methods is to learn a policy that optimizes the selection of actions in a sequential decision making problem \citep{Sutton-98}. A sequential decision making problem is formalized as a Markov decision process or MDP $(\mathcal{S}, \mathcal{A}, P, R)$ where: $\mathcal{S}$ is the state space, $\mathcal{A}$ is the action space, $P$ defines the state transition probability distribution $P(s'|s,a)$ and $R$ defines the reward function $R(s,a,s')$ for taking action $a$ in state $s$ and then transitioning to state $s'$. A policy $\pi$ is a function that maps states into actions. An episode in an MDP consists of a sequence of state, action, reward tuples $(s_t,a_t,r_t)$. Starting with an initial state $s_0$, an episode proceeds according to the policy, state transition distribution and reward function until an absorbing state is reached \citep{Sutton-98}. 

In this work, we focus on two classes of reinforcement learning methods: policy gradient methods and value function methods. Policy gradient methods learn a probabilistic model $\pi_{\theta}$ mapping states into a probability distribution over actions. Value function methods instead learn the value of states or state-action pairs. The domain that we focus on in this work has a factorized state space that includes continuous dimensions, thus we focus on value function methods that can accommodate continuous state variables. We briefly review both classes of methods.

\textbf{Policy Gradient Methods:}
\label{subsubsection policy gradient}
The goal of policy gradient methods is to select the parameters $\theta$ of the policy $\pi_{\theta}$ to maximize the expected return of the policy: $J({\pi}_{\theta}) = \mathop{\mathbb{E}}_{\tau \sim {\pi}_{\theta}} \Big[ R(\tau) \Big]$. Here $R(\tau)$ is the return over a trajectory $\tau$. A trajectory is a sequence of states and actions: $\tau = (s_0, a_0, s_1, a_1,... s_{T-1}, a_{T-1}, s_T)$ where $T$ is the episode length. 

Different policy gradients methods use different definitions of the return $R(\tau)$. In this work we focus on the basic REINFORCE algorithm, which uses a return based on the discounted sum of rewards to go. Policy gradient methods learn the parameters of the policy using a Monte Carlo approximation to the gradient of the expected return function using $M$ sampled trajectories per gradient update \citep{Sutton-99,Williams-92} as shown below where $\gamma$ is the discount rate and $G_t(\tau^{(i)})$ is the reward to go function.
\begin{align}
    \theta_{t+1} &\leftarrow \theta_{t} + \alpha \hat{\nabla} J({\pi}_{\theta})\\
   \hat{\nabla} J({\pi}_{\theta}) &= \frac{1}{M} \sum_{i=0}^{M-1} \sum_{t=0}^{T-1} {\nabla}_{\theta} \log \pi_{\theta} (a_t^{(i)}|s_t^{(i)}) G_t(\tau^{(i)})\\
   G_t(\tau^{(i)}) &= \sum_{k=t}^{T-1}\gamma^{t-k}r_t
\end{align}
One of the interesting properties of REINFORCE as a pure Monte Carlo policy gradient method is that the correctness of the above learning rule and the convergence of the learning algorithm hold in the case where both the policy $\pi_{\theta}$ is modeled using a non-linear function approximator and we only have access to partially observed state vectors $s'_t$ relative to the full MDP state $s_t$. While REINFORCE is known to have high variance, more sophisticated policy gradient methods such as Actor-Critic methods do not have convergence guarantees in continuous state spaces with partially observed state. We also note that while methods like the use of a baseline in the return formulation can also decrease variability, we do not see convergence issues in our experiments when using sufficiently large $M$.

\textbf{Value Function Methods:}
\label{subsubsection value function}
While policy gradient methods aim to directly learn an optimal policy, value function methods such as Q-learning aim to learn the value of state-action pairs and derive a policy by selecting actions that have maximal value in each state \citep{Sutton-98}. In classical Q-learning for discrete state spaces, the state-action value function $Q(s,a)$ is simply a lookup table. More generally, Q-learning can be applied using a function approximator for $Q(s,a)$, which allows Q-learning to be extended to continuous state spaces. For example, the Deep Q Network (DQN) approach uses a deep neural network to approximate $Q(s,a)$ \citep{Mnih-13}.

DQN approaches learn using backpropagation applied to a regression loss $\ell({\delta_t})$ that is a function of the temporal difference error
$\delta_t = r_t + \gamma \cdot \max_{a' \in \mathcal{A}} Q(s_{t+1}, a') - Q(s_t,a_t)$. Fully online learning can be applied after taking each action, but performance can be improved in number of ways including minimizing the loss applied to the temporal difference computed from a batch of examples sampled from a replay buffer and using a second copy of the Q network that is updated more slowly in place of $Q(s_{t+1}, a')$ \citep{deBruin-15, Schaul-16}.

In this work we use the Dueling DQN variant with a replay buffer as an example approach of this class. In the Dueling DQN approach, the Q network is split into two components: a state value function $V(s)$ and a state-dependent advantage function $A(s, a)$. The $Q(s, a)$ value is computed by summing the state value and the advantage value: $Q(s,a) = V(s) + A(s, a)$. The average advantage value $\bar{A}(s) = \frac{1}{|\mathcal{A}|} \sum_{a  \in \mathcal{A}} A(s, a)$ can also be subtracted from the raw advantage value $A(s,a)$ to improve identifiability \citep{Wang-16}.  The model is again learned by minimizing a loss on the temporal difference error. This approach also uses more slowly updated copies of these networks when computing the target $Q(s_{t+1}, a')$ values.

We note that unlike standard Monte Carlo policy gradient methods, Q-learning methods including the Dueling DQN have the ability to learn from trajectories that were not sampled from the current model parameters. This off-policy learning ability allows Q-learning methods to use a replay buffer and provide better sample efficiency. However, Q-learning methods have the significant drawback that their convergence is not guaranteed in a setting where the state is partially observed and state-action values are represented using non-linear function approximators, including neural networks. 

%Ben: Description of properties of RL methods under partial observability
  % Lack of convergence of value function methods with non-linear function approximation
  % The belief MDP interpretation of martial observability
  % The convergence of policy gradient methods under partial observability

\subsection{RL for JITAIs}
\label{subsec:rl-jitais}
% Ben: Description of Susan's recent work, literature review on other RL work for JITAI and adaptive interventions. Maybe also Daniel's work on control system methods

Prior work on RL methods for JITAIs has largely focused on contextual bandit methods \citep{paredes2014poptherapy, rabbi2015mybehavior, tewari2017ads, yom2017encouraging}. These methods aim to select actions that maximize the immediate expected reward, thus discounting longer term effects of actions. However, adaptive health intervention domains can have significant long term and delayed effects. To address this challenge \cite{liao2020personalized} develop an extended bandit-like algorithm that uses a model-based proxy reward to imitate the longer term effect of actions. \cite{gonul2021reinforcement} propose an RL method that uses modified eligibility traces that aim to credit intervention components that the participant actually engaged with. The core RL algorithm used is based on Q-learning, but assumes that discrete states are provided by an auxiliary state classifier. 

While both \cite{liao2020personalized} and \cite{gonul2021reinforcement} represent improvements over contextual bandit methods in terms of their ability to model longer term effects of actions, both approaches condition on context variables as if they are known without uncertainty, which is the specific issue we study in this work. Further, through the use of the auxiliary state classifier, \cite{gonul2021reinforcement} avoid issues that arise when composing Q-learning methods with function approximation under partial observability, which we also address directly. 

Finally, we note that \cite{liao2020personalized} articulate multiple important practical challenges with the deployment of RL methods for JITAIs including the need for methods that can learn quickly from limited interactions with single individuals. In this work our primary goal is to quantify the fundamental limits imposed by context inference error and partial observability. As a result, we do not consider restrictions on the number of simulated interactions with a user or restriction on the number of episodes of training. Our results should be interpreted as establishing upper bounds on the performance achievable by methods that impose further constraints.

\begin{table}[t]
  \centering
  \caption{Actions Values}
  \label{tab:actions}
  \begin{tabular}{cl}
    \toprule
    \bfseries Action Value & \bfseries Description \\
    \midrule
    $a=0$   & do not send a message \\
    $a=1$   & send a non tailored message\\
    $a=2$   & send a message tailored to context $0$ \\
    $a=3$   & send a message tailored to context $1$\\
    \bottomrule
  \end{tabular}
\end{table}

\section{Methods}
\label{sec:methods}

In this section we describe the physical activity JITAI simulation environment that we use in this work as well as the context error and partial observability conditions that we study. We also describe in detail the reinforcement learning agents used in our experiments.

\subsection{Physical Activity JITAI Simulation Environment}
%Karine: draft the description of the simulation environment
%Define the state variables: C_t, H_t, D_t, P_t, L_t
%Let us use C_t for true context, P_t for context probability, and L_t for most likely context
%Define the actions.
%Define the transition functions. Give the probabilistic versions.
%Provide a table defining all of the variables (\rho, etc). Give their symbols,
%descriptions, and range of values used in experiments.

We design a JITAI simulation environment taking inspiration from recent work in the area of contextualized messaging based intervention studies for promoting walking as a form of physical activity \citep{hardeman2019systematic, spruijt2022advancing}. Below we describe the state, action, and dynamics of the physical activity JITAI simulation.

\textbf{State and Actions:}
A contextualized messaging intervention leverages a pool of messages that aim to provide support in different contexts. The choice of whether and what type of message to send at each time step depends on the individual's context $c_t$. We select stressed/not stressed as an example binary context variable in our simulation. As discussed in the previous section, such context variables are often derived from sensor-based inferences \citep{hovsepian2015cstress}.  To reflect the fact that the true context is not known to the reinforcement learning agent, we use $\mathbf{p}_t$ to denote an inferred probability distribution over the context, and $l_t$ to represent the most likely context value according to $\mathbf{p}_t$.

In addition to the stressed/not stressed context variable, we model two additional psychological state variables: habituation $h_t$ and disengagement risk $d_t$. Intuitively, habituation models the extent to which the effect of the intervention is attenuated through prior exposure to the intervention. Disengagement risk facilitates modeling a common problem with adaptive interventions: in response to factors such as perceived lack of utility, intervention participants sometimes completely abandon the use of an intervention. We  discuss the dynamics of these variables in the next section.

\begin{table}[t]
    \centering
    \caption{Simulation Variables}
    \label{tab:env state}
    \begin{tabular}{cll}
      \toprule
      \bfseries Variable & \bfseries Description  & \bfseries Values\\
      \midrule
            $c_t$     & true context                  & $\{0,1\}$ \\
            $\mathbf{p}_t$ & context probabilities    &  $\Delta^1$\\
            $l_t$     & most likely context           & \{0,1\}\\
            $d_t$     & disengagement risk level      & $[0,1]$\\
            $h_t$     & habituation level             & $[0,1]$\\
            $s_t$     & number of steps               & $\mathbb{N}$\\
      \bottomrule
    \end{tabular}
\end{table}

We summarize the variables in the simulation and their value ranges in Table \ref{tab:env state} (note that $\Delta^1$ indicate the probability simplex for a binary variable). The simulation includes a total of four actions as summarized in Table \ref{tab:actions}. Action $a=0$ is the null action where no message is sent. Action $a=1$ corresponds to sending a non-context tailored message. Actions $a=3$ and $a=4$ correspond to sending messages tailored to context 0 and 1 respectively. Note that based on the numerical context and action values, $a_t=c_t+2$ corresponds to the selection of a message that is tailored for the correct context. In response to taking an action in a given state at time $t$, we observe a reward in the form of a step count $s_t$.

\textbf{Dynamics:}
We focus on simulating the dynamics of habituation and disengagement and how they relate to the effect of the intervention components. 
We model habituation as increasing with each message sent up to an upper limit and decaying towards zero when messages are not sent. 
We model disengagement risk as increasing only when incorrectly contextualized messages are sent and decaying towards zero only when uncontextualized or correctly contextualized messages are sent. We provide the update equations for these state variables below. The parameters of the update equations are described in Table \ref{tab:env config}.
\begin{align*}
h_{t+1} &=   \begin{cases}
                (1-\delta_h) \cdot  h_t             &\text{if~} a_t = 0\\
                \text{min}(1, h_t + \epsilon_h)     & \text{otherwise}\\
            \end{cases}\\
d_{t+1} &=   \begin{cases}
                d_t                                 &\text{if~} a_t = 0\\
                (1-\delta_d) \cdot  d_t             &\text{if~} a_t = 1 ~\text{or}~ a_t=c_t+2\\
                \text{min}(1, d_t + \epsilon_d)     &\text{otherwise}
            \end{cases}
\end{align*}

\begin{table}[t]
    \centering
    \caption{Environment Parameter Settings.}
    \label{tab:env config}
    \begin{tabular}{cll}
      \toprule
      \bfseries Parameter & \bfseries Description  & \bfseries Value\\
      \midrule
            $\delta_h$     & habituation decay           & 0.1 \\
            $\epsilon_h$   & habituation increment       & 0.05 \\
            $\delta_d$     & disengagement decay         & 0.1  \\
            $\epsilon_d$   & disengagement increment     & 0.4 \\
            $\rho_1$       & $a_t=1$ base reward         & 50. \\
            $\rho_2$       & $a_t=c_t+2$ base reward     & 200. \\
            $\sigma$       & feature uncertainty         & $\{0.4,..., 2\}$ \\
      \bottomrule
    \end{tabular}
      \vspace{1em}
\end{table}

\begin{figure}[t]
    \centering
    \includegraphics[width=2.5in]{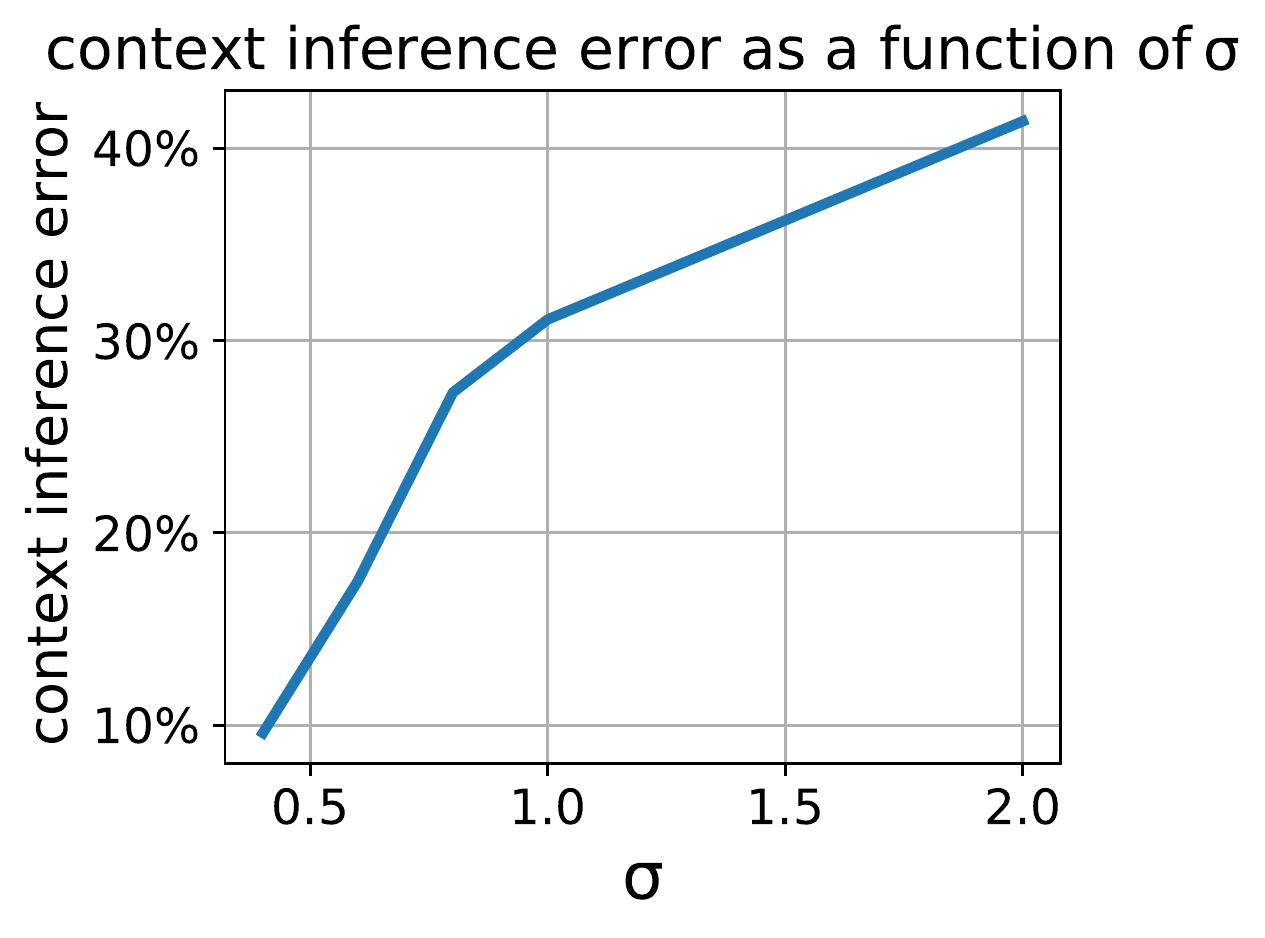}
    \caption{Context inference error as a function of $\sigma$.}
    \label{fig:uncertainty_context_inferred_error_vs_sigma}
\end{figure}

We model the reward in terms of the surplus step count generated beyond a potentially
context dependent baseline level $\mu_c$. We model incorrectly contextualized messages
and not sending a message as generating zero surplus reward. We model uncontextualized actions
and correctly contextualized actions as providing base surplus rewards $\rho_1$ and $\rho_2$ 
that are attenuated by the habituation level $h_t$. Specifically, as the  habituation level increases,
the fraction of the base reward that is realized decreases. While increasing disengagement risk does not
have an immediate effect on reward, if the disengagement risk reaches the value $1$, we simulate the occurrence of
a disengagement event that terminates the episode. This delayed effect can have a significant
impact on total reward over an episode. The maximum length of an episode is set to $50$ time steps.
\begin{align*}
s_{t+1} &=   \begin{cases}
                \mu_{c_t}    + (1-h_{t+1}) \cdot  \rho_1  &\text{if~} a_t = 1\\
                \mu_{c_t}    + (1-h_{t+1}) \cdot  \rho_2  &\text{if~} a_t = c_t+2\\
                \mu_{c_t}    & \text{otherwise}
            \end{cases}
\end{align*}
We model the true context as a purely random Bernoulli process. At each time step we sample $c_t \sim\mbox{Bernoulli}(0.5)$.
To model a sensor-derived inference for $c_t$, we follow a two step process. We sample a normally distributed
context-dependent scalar feature $x_t \sim \mathcal{N}(c_t, \sigma^2)$ where $\sigma$ models the uncertainty in the
feature given the context. We next compute the context probability distribution $\mathbf{p}_t$ given the sampled feature 
value $x_t$ as $p_{ct}=P(C_t=c | x_t)$ simulating the application of probabilistic context classifier. Finally, we set the most likely context to $l_t = \argmax_c \; p_{ct}$. We vary the feature noise standard deviation parameter $\sigma$  from $0.4$ to $2$. This generates context inference errors varying from $10\%$ to $41\%$.  Figure \ref{fig:uncertainty_context_inferred_error_vs_sigma} shows the effect of the feature noise standard deviation parameter $\sigma$ on the context inference error rate.

%-----------------------------------------------------------------------

\subsection{Context Inference and Partial Observability Conditions}
%Karine: Describe the different observation functions used in the experiments.
%Scenario 1: C_t, H_t, D_t all observed
%Scenario 2: L_t, H_t, D_t observed
%Scenario 3: P_t, H_t, D_t observed
%Scenario 4: C_t, t observed
%Scenario 5: L_t, t observed
%Scenario 6: P_t, t observed

%

We consider six different scenarios in terms of the observations that are provided to the RL agent during learning. 
The full state consists of the triple $(c_t,h_t,d_t)$. We consider the case where $c_t$ is not directly observed and we
instead provide the agent with either the most likely inferred context $l_t$ as an input, and the case where $c_t$
is not directly observed and we provide the agent with information about the inferred probability distribution over the context variable $\mathbf{p}_t$ as input. Specifically, since the distribution $\mathbf{p}_t$ is over a binary variable, we supply $p_{0t}$ (the probability that the context is $0$ as the feature. 
Further, we consider the case where the
state variables $h_t$ and $d_t$ are both observed and the case where neither is observed. When 
$h_t$ and $d_t$ are not observed we augment the state with a time indicator variable $i_t$. In our experiments we use
a time indicator variable $i_t=\mbox{mod}(t,k)$. This choice enables the agent to take different actions based
on a cyclic notion of time within an episode. We experimented with different values of $k$ and found little difference between different small values of $k$. We use $k=2$ in our experiments. 

In our experiments, the scenarios described above are labeled as follows:
C-H-D: $c_t$, $h_t$, $d_t$ observed.
L-H-D: $l_t$, $h_t$, $d_t$ observed.
P-H-D: $p_t, h_t, d_t$ observed.
C-T: $c_t, i_t$ observed. 
L-T: $l_t, i_t$ observed.
P-T: $p_t, i_t$ observed.

We expect agents learned using the C-H-D observation set to perform the best as these agents have access to the full MDP state space. We hypothesize that as the feature noise increases, the P-H-D observation set will perform better than the L-H-D feature set as access to the context inference probability distribution provides the agent with strictly more information than the most likely context. 
Finally, we hypothesize a loss in performance in the scenarios where the habituation and disengagement variables can not be observed, which is a more realistic scenario as these variables can not be passively sensed and are problematic to obtain in practice even via direct self report.

%-------------------------------------------------------------

\subsection{Reinforcement Learning Agents}
\label{Reinforcement Learning Agents}

%Karine: Describe the details for the Q-learning agent. What is the Q-Network. What version of the learning problem is being used. What are the hyper-parameters and how were they set. What optimization method was used. What re-play/batching and other details. 

%Karine: Describe the Reinforce agent. What is the policy etwork. What are the hyper-parameters and how were they set. What optimization method was used. 

\begin{figure*}[t]
    \centering
     \begin{subfigure}[b]{0.24\linewidth}
             \includegraphics[width=\linewidth]{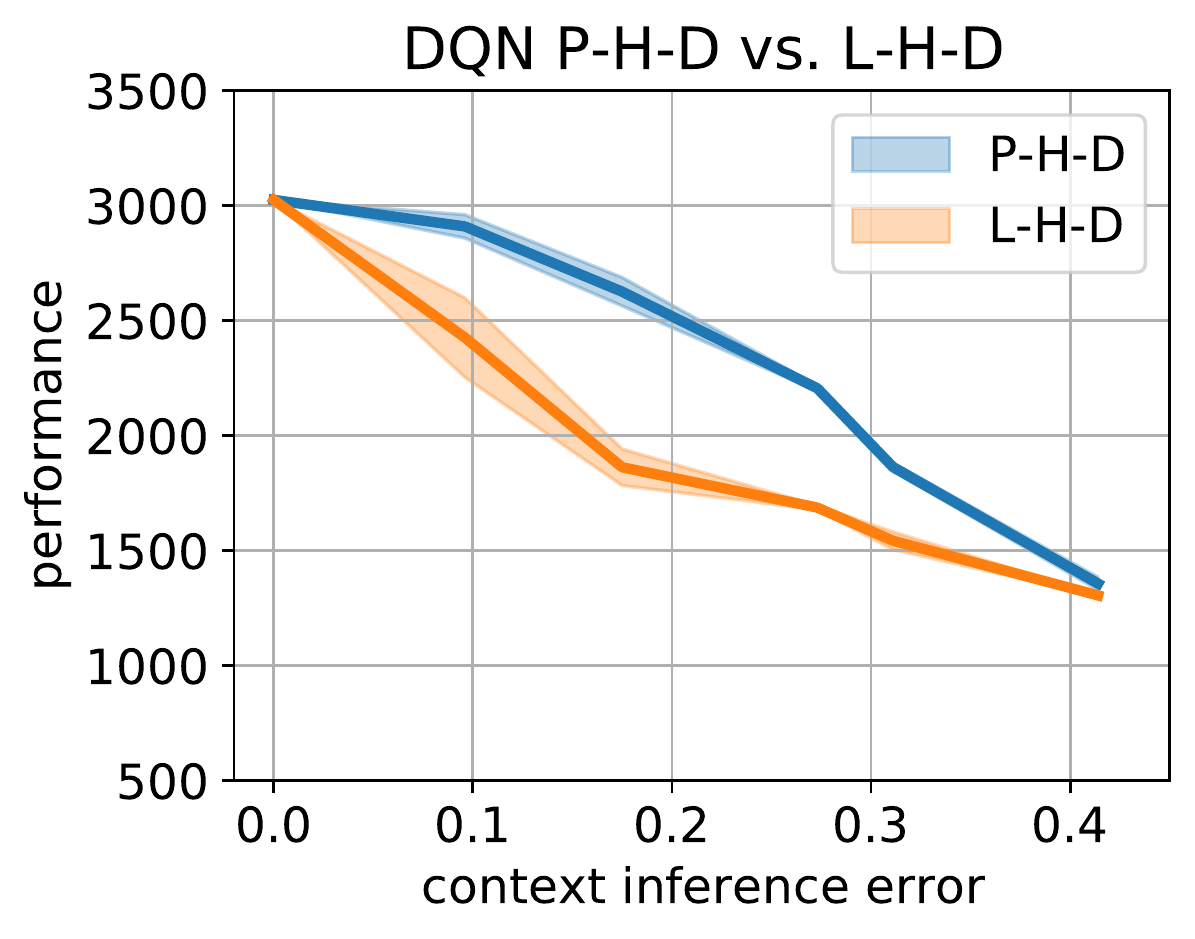}
             \caption{}
            \label{fig:perf_vs_error_dqn}
     \end{subfigure}
     \begin{subfigure}[b]{0.24\linewidth}
            \includegraphics[width=\linewidth]{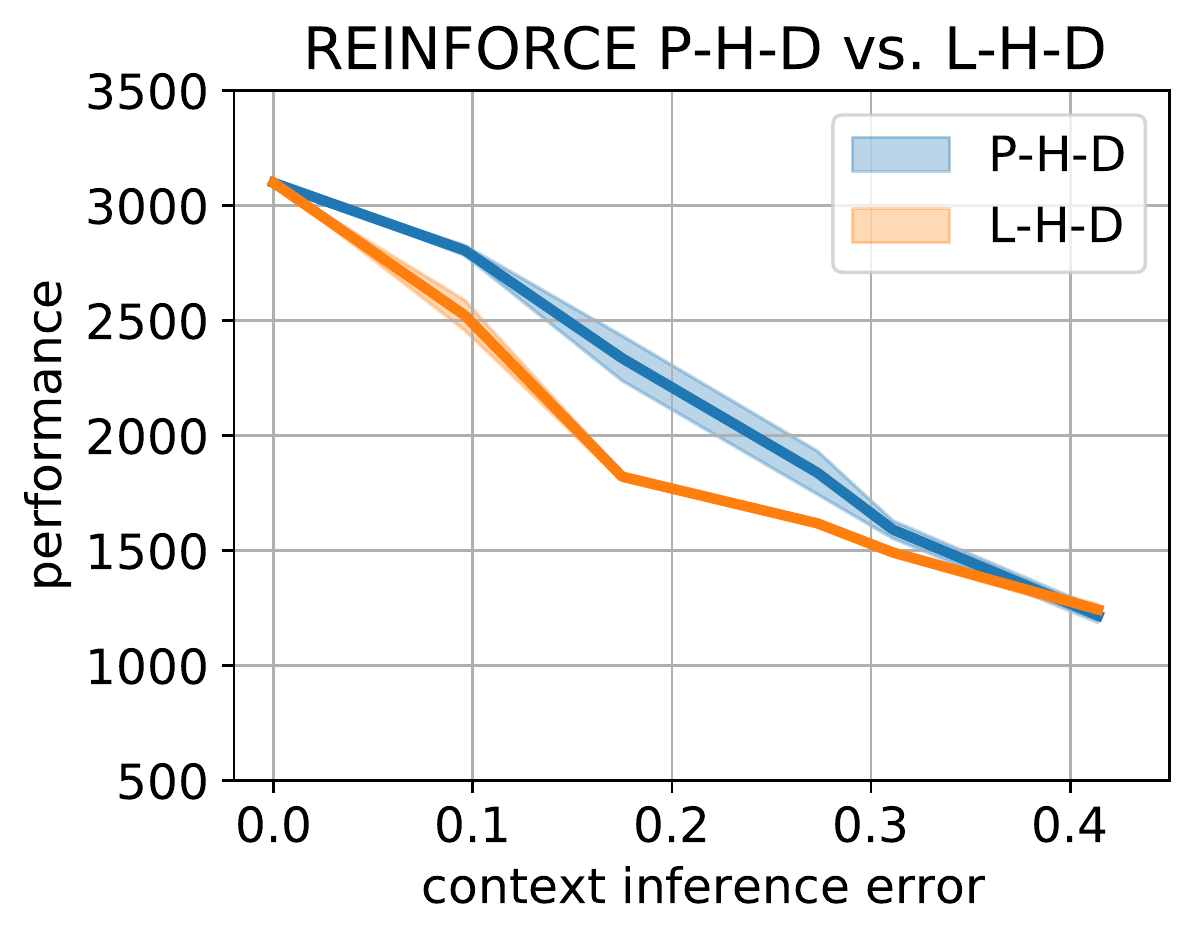}
            \caption{}\label{fig:perf_vs_error_reinforce}
     \end{subfigure}
     \begin{subfigure}[b]{0.24\linewidth}
            \includegraphics[width=\linewidth]{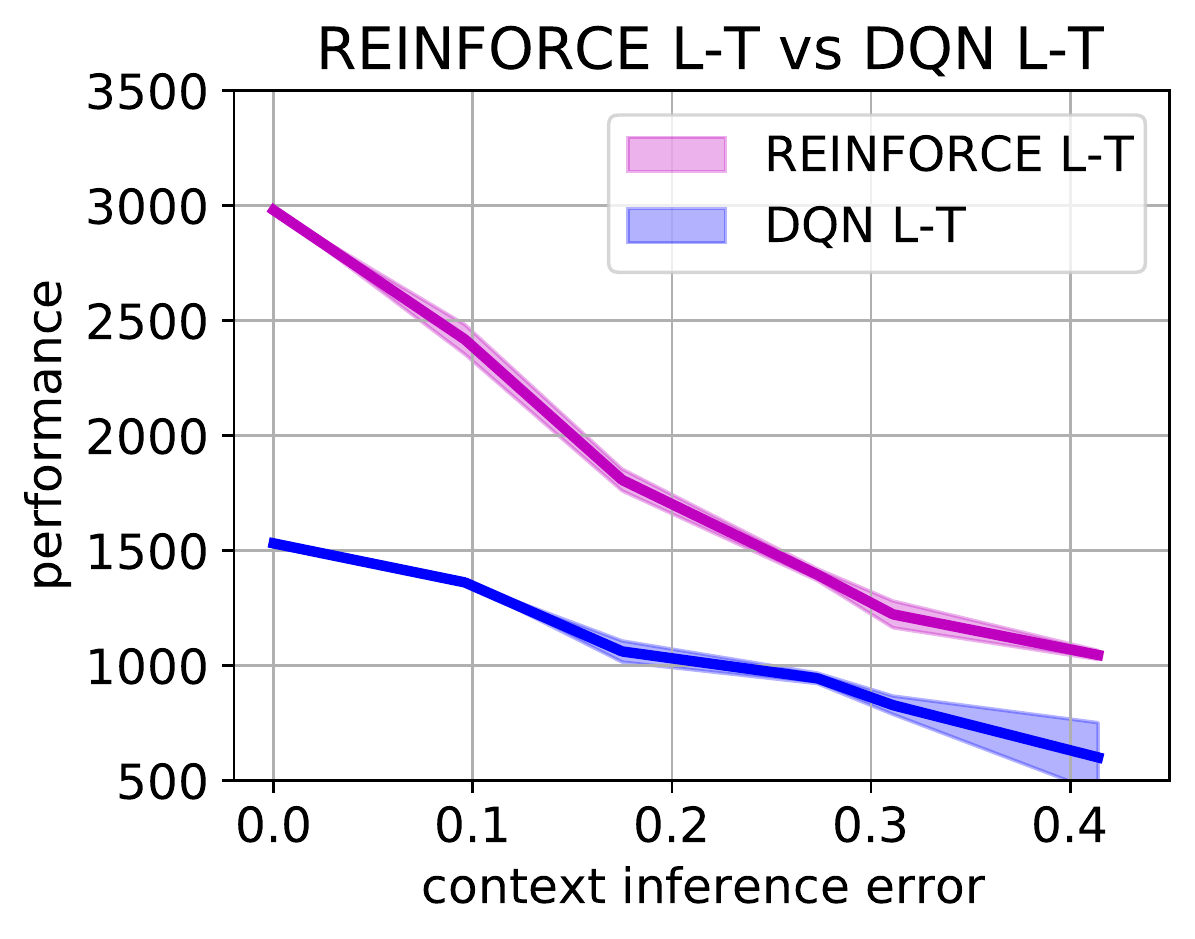}
            \caption{}\label{fig:perf_reinforce_dqn_ltfn_vs_error}
     \end{subfigure}
     \begin{subfigure}[b]{0.24\linewidth}
        \includegraphics[width=\linewidth]{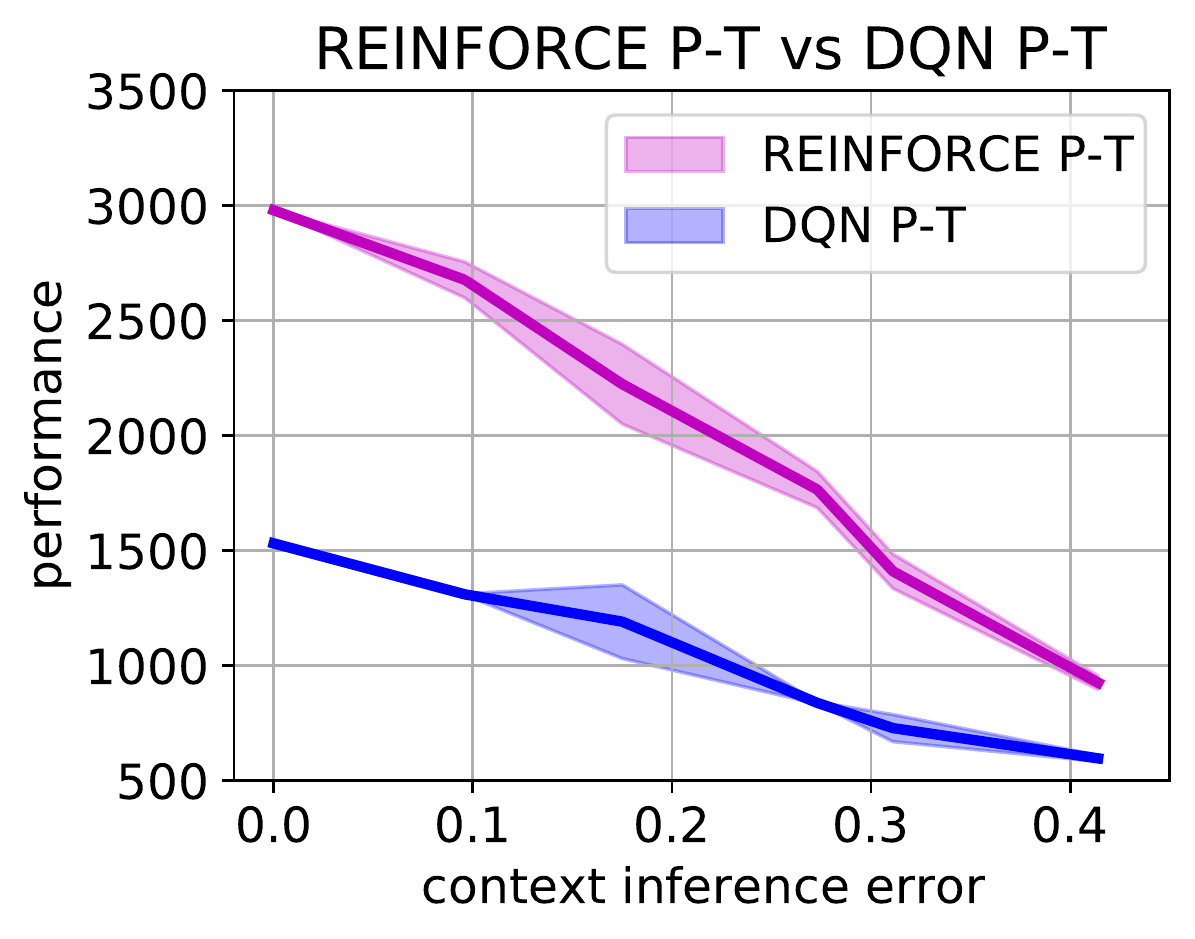}
        \caption{}\label{fig:perf_reinforce_dqn_ptfn_vs_error}
     \end{subfigure}
    \caption{(a) Effect of learning with most likely context and context probabilities for DQN. (b) Effect of learning with most likely context and context probabilities for REINFORCE.
    (c) Effect of learning with most likely contexts and partial observability for REINFORCE and DQN.
    (d) Effect of learning with context probabilities and partial observability for REINFORCE and DQN.}
\end{figure*}

In our experiments, we compare a policy gradient method to a value function method. For the value function method, we select the Dueling DQN method. We use a multilayer perceptron with two hidden layers, for both the state value and advantage functions. We perform a hyper-parameter search over hidden layers sizes $[32, 64, 128, 256]$, batch sizes $[16, 32, 64]$ Adam optimizer learning rates from $1\text{e-}6$ to $1\text{e-}2$, and epsilon greedy exploration rate decrements from $1\text{e-}6$ to $1\text{e-}3$. We report results using with $128$ neurons on each hidden layer, Adam optimizer learning rate $lr = 5\text{e-}4$, epsilon linear decrement $\delta_{\epsilon} = 0.001$, decaying $\epsilon$ from $1$ to $0.01$, batch size $64$. The target Q network parameters are replaced every $K = 1000$ steps. The number of episodes used to learn the model is $1000$.

For the REINFORCE policy network, we use a multilayer perceptron with one hidden layer. We perform hyper-parameter search over hidden layer sizes $[32, 64, 128, 256]$, and Adam optimizer learning rates from $1\text{e-}6$ to $1\text{e-}2$. We report results using $128$ neurons, and Adam optimizer learning rate $lr = 6\text{e-}4$. We set the number of trajectory samples per gradient step to $M = 50$ and the number of episodes used for learning to $15,000$.

\begin{figure*}[t]
    \centering
    \includegraphics[height=3cm]{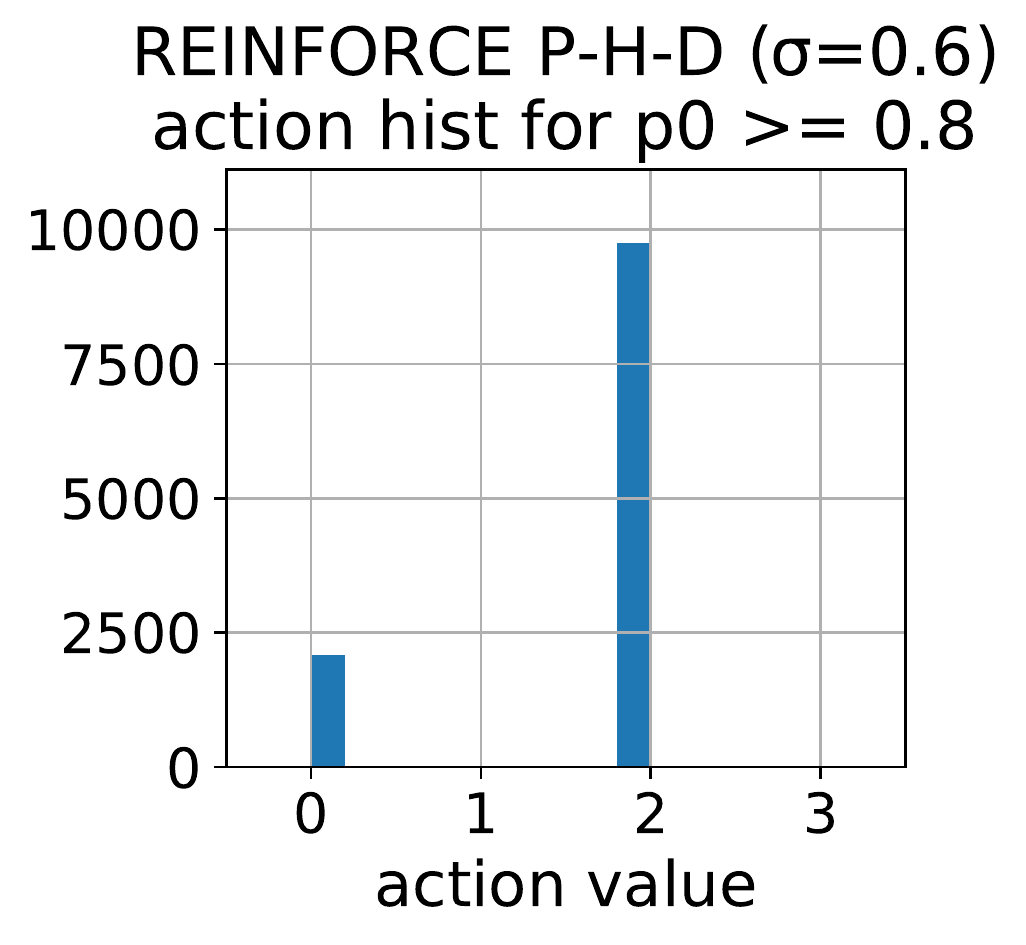}
    \includegraphics[height=3cm]{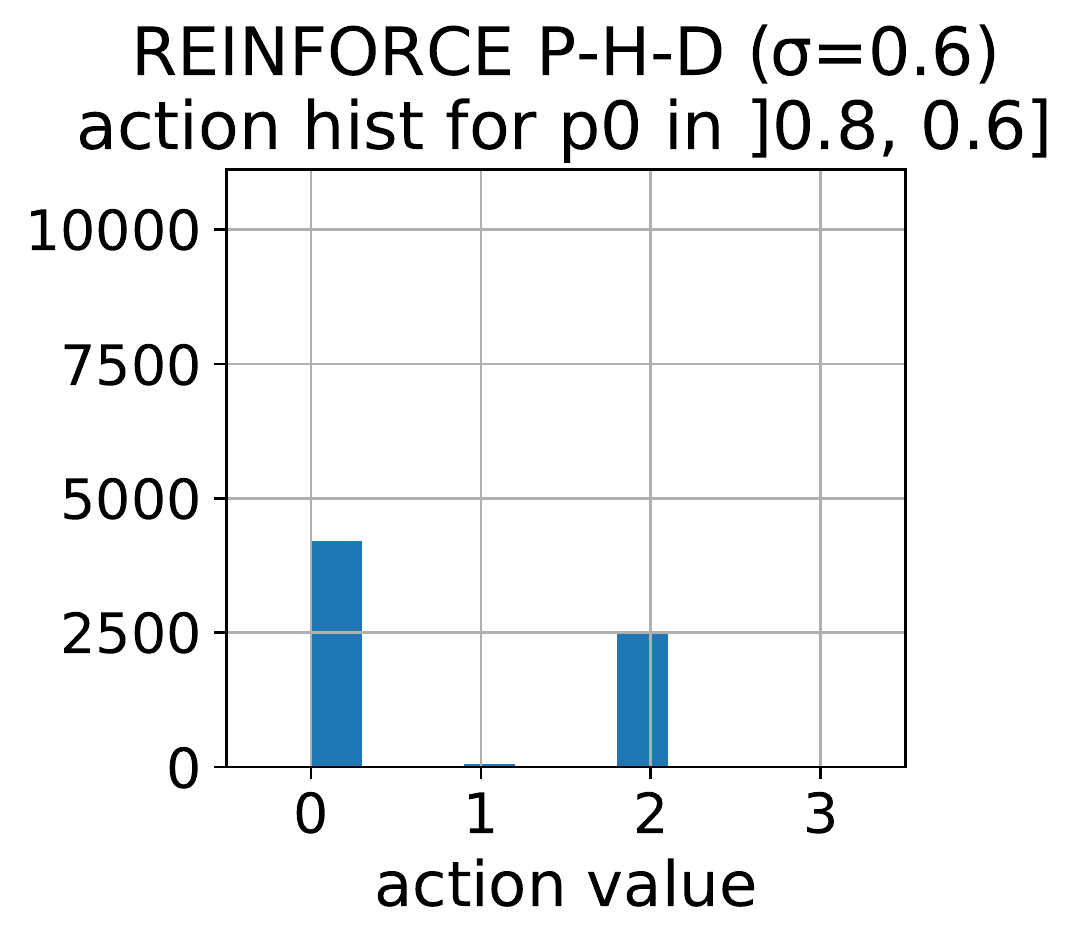}
    \includegraphics[height=3cm]{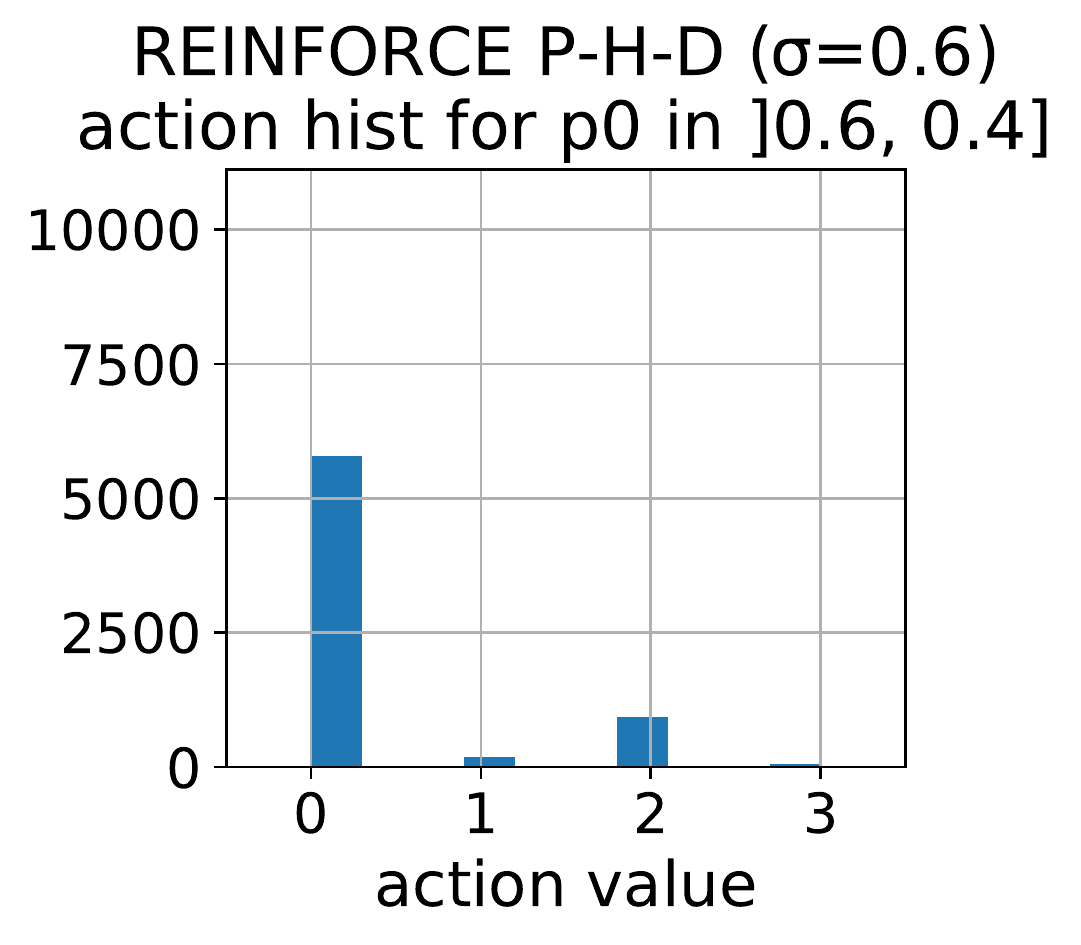}
    \includegraphics[height=3cm]{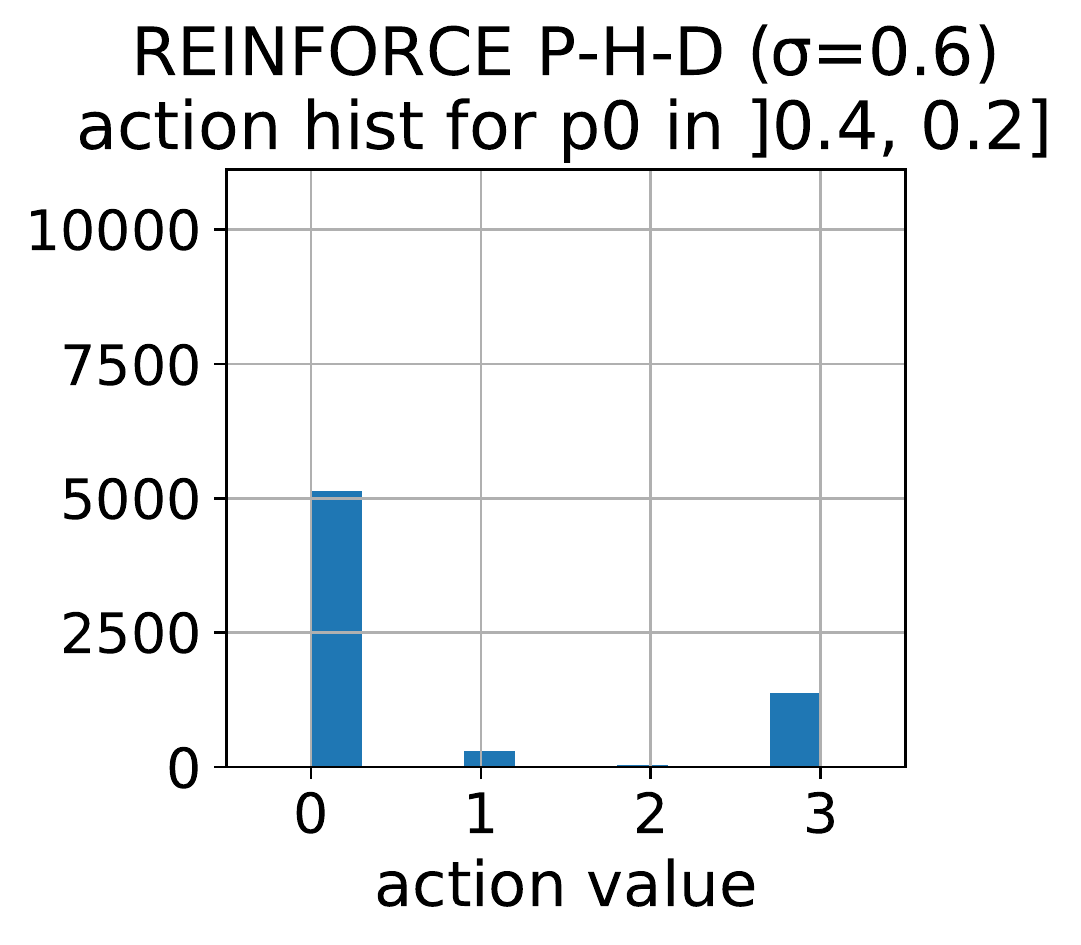}
    \includegraphics[height=3cm]{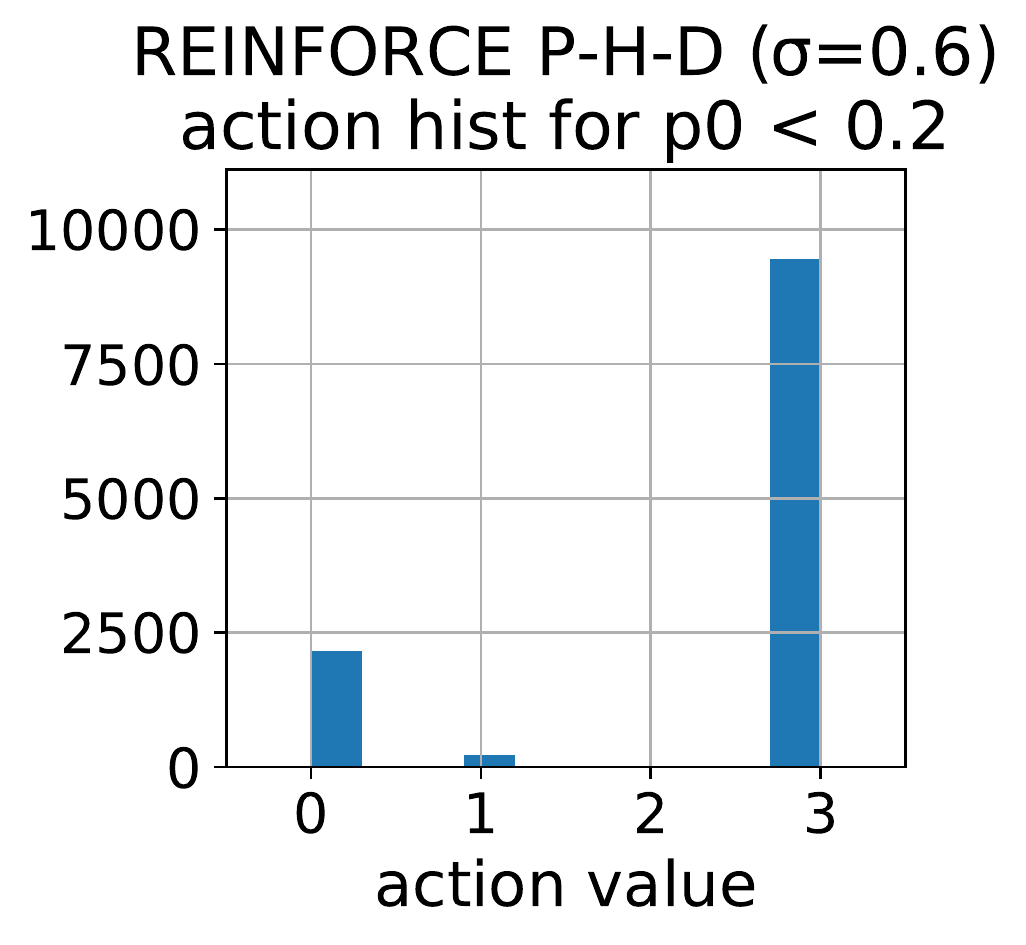}\\
    \centering
    \includegraphics[height=3cm]{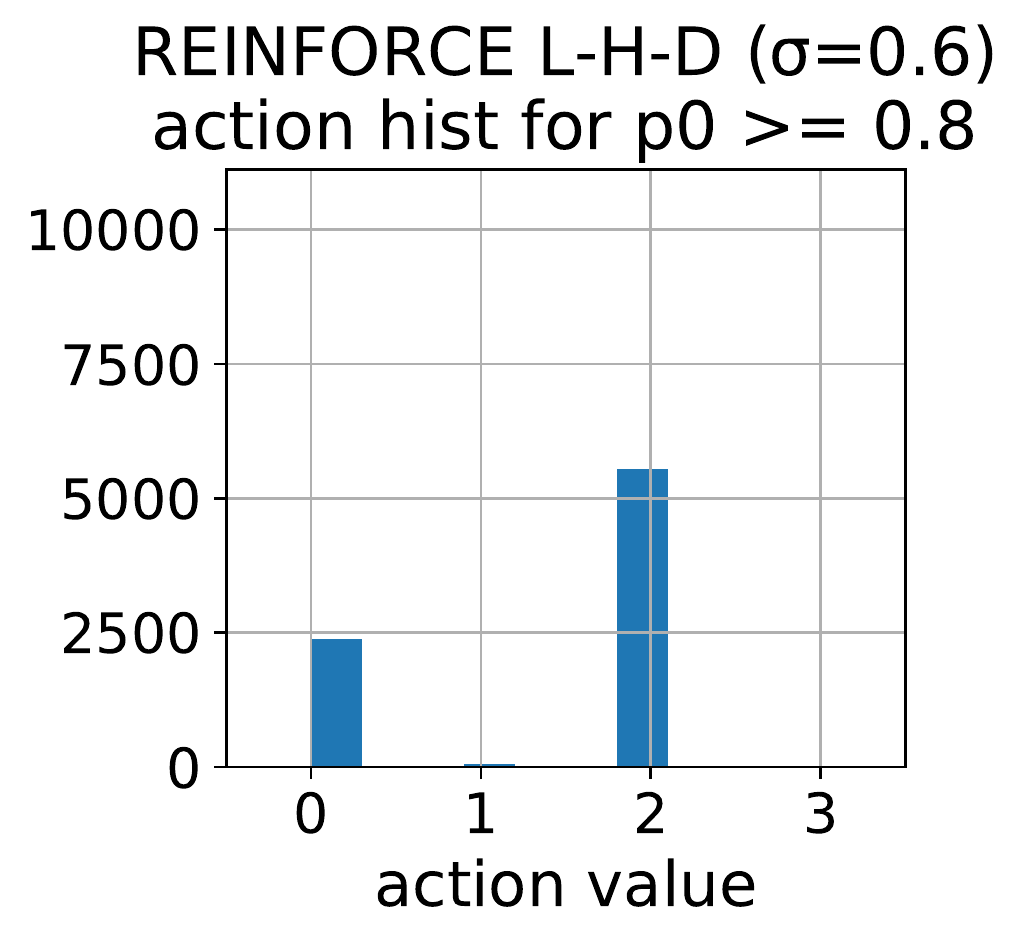}
    \includegraphics[height=3cm]{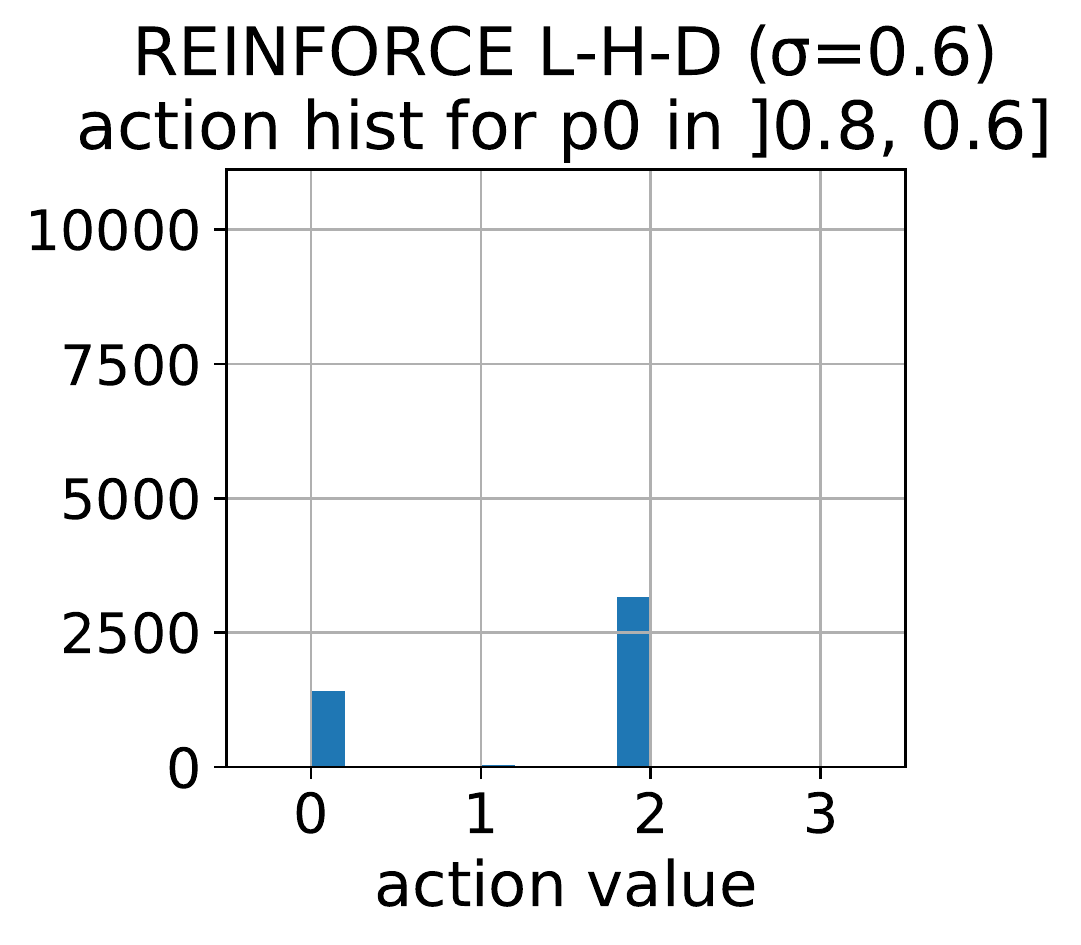}
    \includegraphics[height=3cm]{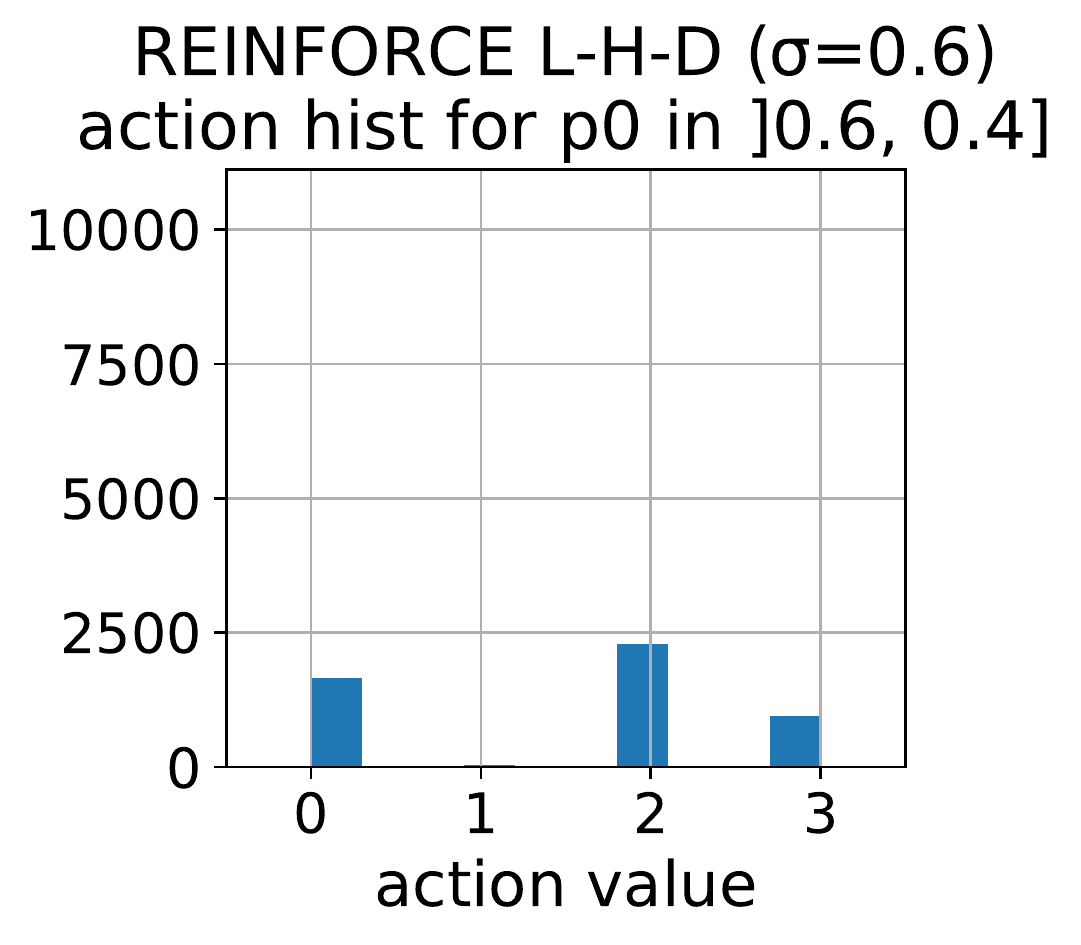}
    \includegraphics[height=3cm]{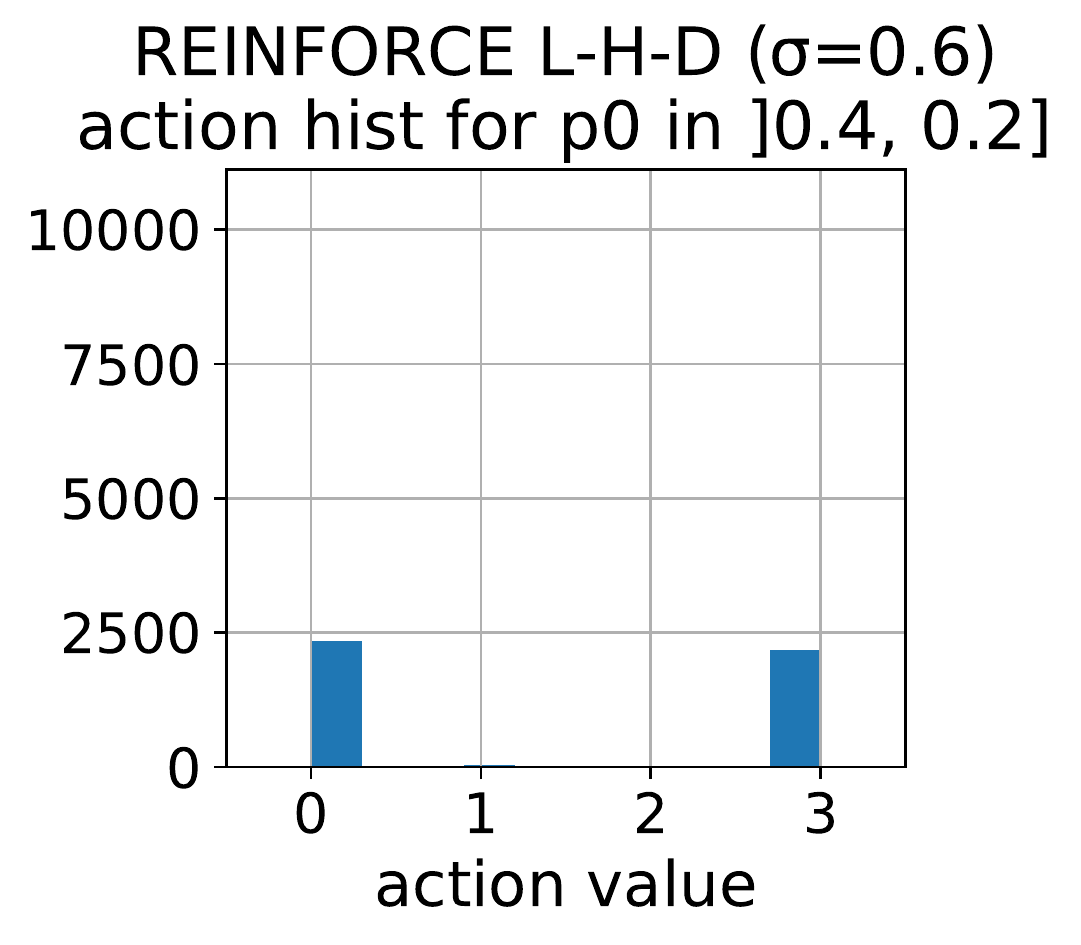}
    \includegraphics[height=3cm]{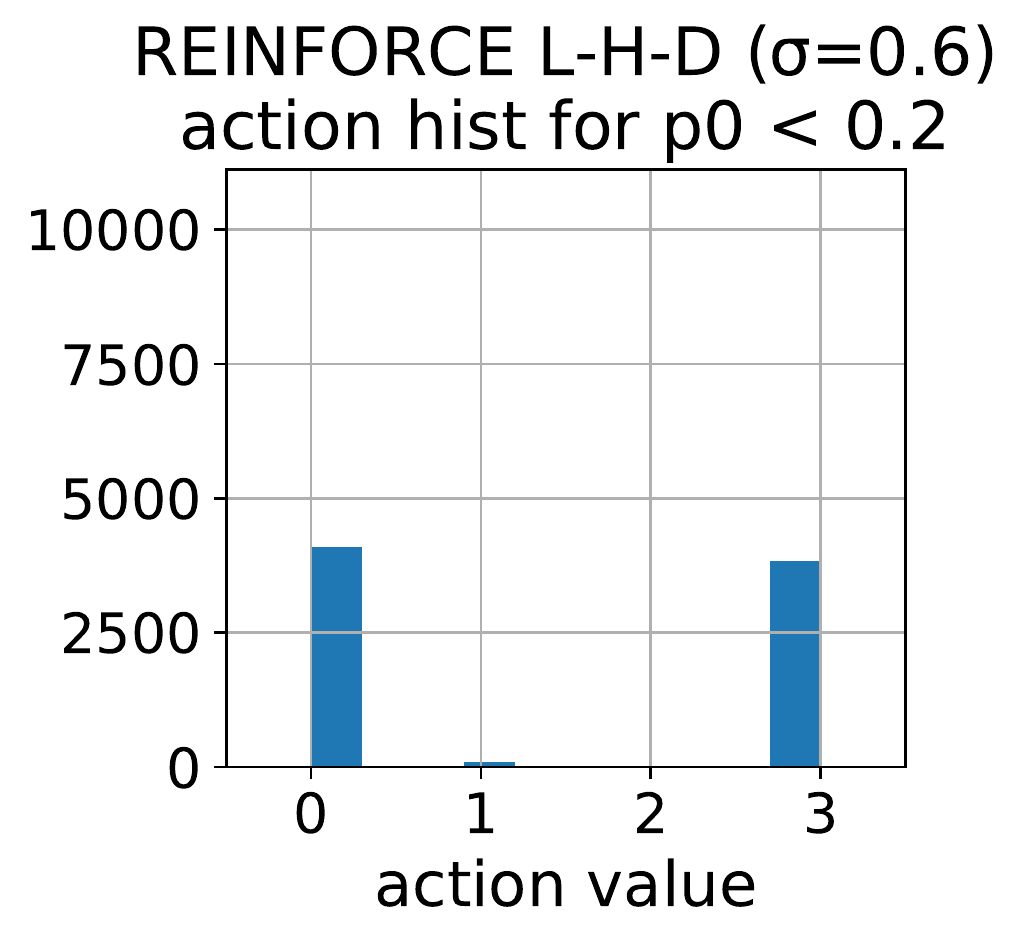}
    \caption{The top row of plots shows
    the distribution of actions selected by REINFORCE when given access to context probabilities. The bottom row of plots shows
    the distribution of actions selected by REINFORCE when given access only to the inferred most likely context. }
    \label{fig:action_distributions}
\end{figure*}

\section{Experiments and Results}\label{sec:experiments}
In this section we present experiments and results using the physical activity JITAI simulation domain and the reinforcement learning agents and scenarios introduced in the previous section. 
We repeat each experiment 3 times with different random seeds. All experiments use a reward discount rate of $\gamma = 0.99$. In all the experiments and for all random seeds, we first learn a policy and then compute the performance of the policy using the average over $1000$ test episodes of the per-episode non-discounted total reward. We report the average performance over three seeds as well as the standard deviation of the performance over three seeds.

%-------------------------------------------

\textbf{The Effect of Learning with Most Likely Contexts:} We begin by quantifying the impact of learning policies given the most likely context $l_t$ instead of the true context $c_t$ under the assumption that the habituation and disengagement variables are fully observed. In this experiment we vary the value of feature uncertainty parameter $\sigma$ from $0$ to $2$ resulting in variation in context inference error from 0\% to approximately 40\%. 
As described in the previous section, we repeat this experiment three times for three random seeds for both DQN and REINFORCE and report performance in terms of average per-episode total reward. The results are shown as the orange lines in figures \ref{fig:perf_vs_error_dqn} and \ref{fig:perf_vs_error_reinforce} for the DQN and REINFORCE agents. As we can see, the best performing policies are obtained when the context inference error rate is $0$ so that $l_t=c_t$. As the context inference error rate increases, the performance of both the DQN and REINFORCE agents drops quickly. We can see that at a context inference rate of 40\%, both agents experience a drop in reward due to using most likely contexts, of approximately 50\% relative to using true contexts.

%-------------------------------------------

\textbf{The Effect of Learning with Context Probabilities:} We next quantify the impact of learning policies given access to context inference probabilities $\mathbf{p}_t$ instead of the true context $c_t$ under the assumption that the habituation and disengagement variables are fully observed. We contrast access to context inference probabilities with access only to most likely inferred contexts. We use the same experimental procedure as for the previous experiment. The results are shown as the blue lines in figures \ref{fig:perf_vs_error_dqn} and \ref{fig:perf_vs_error_reinforce} for the DQN and REINFORCE agents. As expected, the best performing policies are again obtained when the feature uncertainty level is $\sigma=0$ and the context inference error rate is $0$ so that $p_t$ effectively carries the same information as $c_t$. As the context inference error rate increases, the performance of both the DQN and REINFORCE agents using $\mathbf{p}_t$ again decreases. 

However, as we can see from the figures, the performance of the agents with access to $\mathbf{p}_t$ clearly dominates the performance of agent with access to $l_t$ until the context inference error rate approaches the maximum value considered. At a context inference error rate of 0.17, both agents using $\mathbf{p}_t$ achieve an increase of performance of more than 500 steps relative to the agents using only $l_t$. 

\begin{table}[t]
  \centering
  \caption{Unpaired t-tests on performance for scenarios P-H-D vs. L-H-D, for different error rates, for both agents. Effect is the difference of the average returns.}
  \label{tab:unpaired t-tests}
  \begin{tabular}{cccc}
    \toprule
    \bfseries P-H-D vs. L-H-D & \bfseries Error Rate & \bfseries Effect & \bfseries p-value \\
    \midrule
    DQN  & 10\%  & 483.02 & 0.01930 \\
    DQN  & 17\%  & 763.36 & 0.00043 \\
    DQN  & 27\%  & 518.49 & 0.00000 \\
    DQN  & 31\%  & 320.38 & 0.00045 \\
    DQN  & 41\%  & 51.42  & 0.06041 \\
    \midrule
    REINFORCE  & 10\% & 282.42 & 0.00442 \\
    REINFORCE  & 17\% & 514.57 & 0.00183 \\
    REINFORCE  & 27\% & 220.28 & 0.03019 \\
    REINFORCE  & 31\% & 99.38  & 0.02693 \\
    REINFORCE  & 41\% & -26.18 & 0.37660 \\
    \bottomrule
  \end{tabular}
\end{table}

To formally assess the differences between agents with access to $\mathbf{p}_t$ and $l_t$, we perform unpaired t-tests over the three repetitions for each context inference error rate. The results are shown in Table \ref{tab:unpaired t-tests}. A p-value $<0.05$ indicates a statistically significant difference. The unpaired t-tests confirm that up to a context error rate of approximately 30\%, access to $\mathbf{p}_t$ results in statistically significant improvements in total reward compared to access to $l_t$.

We provide more insight into the effect of access to context inference probabilities compared to most likely context inferences in Figure \ref{fig:action_distributions}. The top row of plots shows the distribution of actions selected by REINFORCE when given access to context probabilities. The bottom row of plots shows the distribution of actions selected by REINFORCE when given access only to the inferred most likely context. Each plot in each row corresponds to the distribution of actions in a specific range of context inference probabilities. All results are for a context inference error rate of 17\%. 

As we can see, when given access to context inference probabilities, REINFORCE increasingly avoids taking the contextualized message actions 2 and 3 as the context uncertainty increases, instead preferring to take action 0. When the context inference uncertainty is low, it takes contextualized actions most of the time. By contrast, when given only the most likely inferred context as input, REINFORCE takes a larger proportion of actions 2 and 3 when the context is uncertain, resulting in a higher rate of disengagement events.  Figure 1 in the supplemental material shows similar results for the DQN agent.  

\begin{figure*}[t]
    \centering
    \includegraphics[width=5cm]{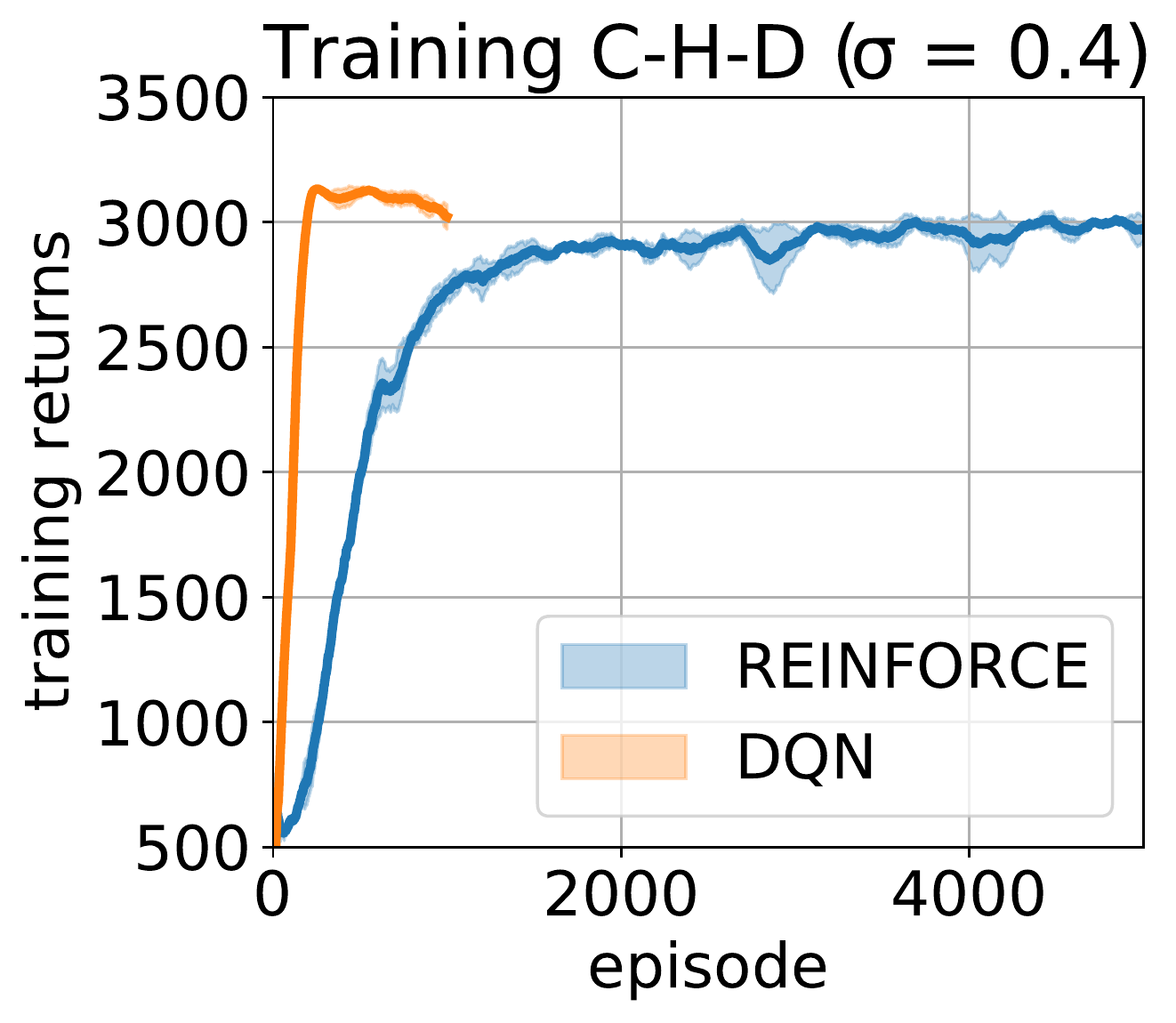}
    \includegraphics[width=5cm]{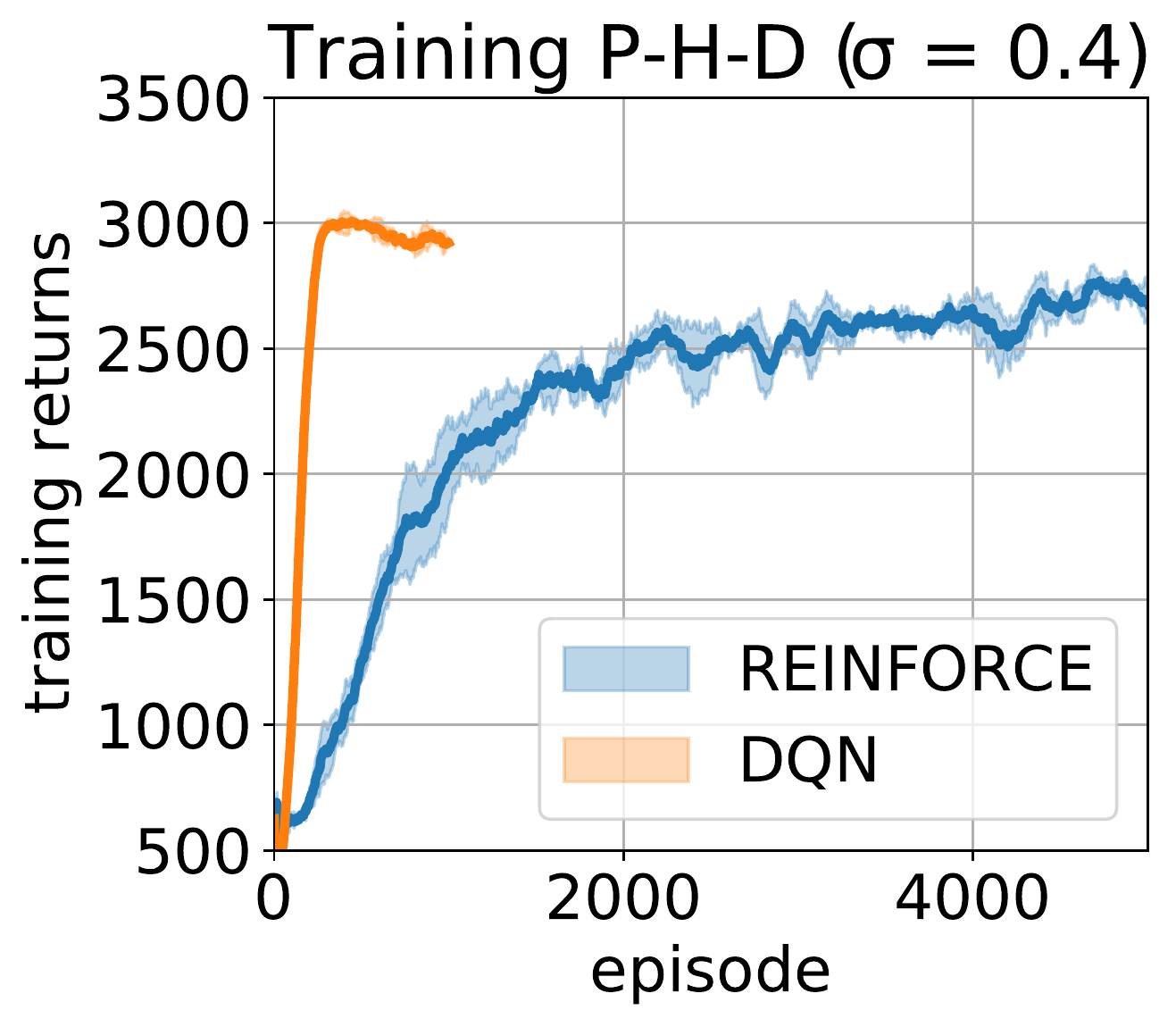}
    \includegraphics[width=5cm]{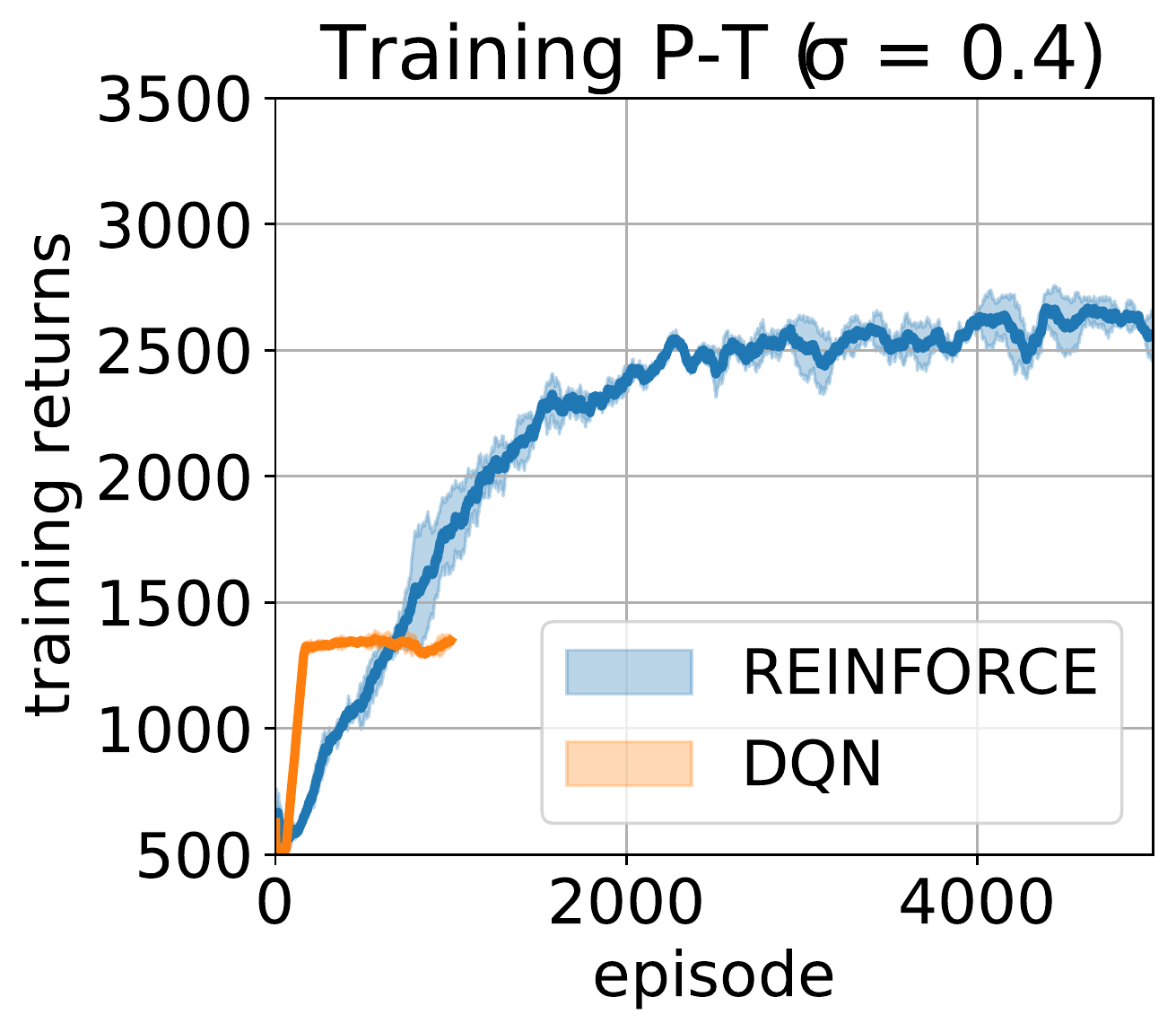}
    \caption{Learning curves of DQN and REINFORCE.}
    \label{fig:train_dqn_vs_reinforce}
\end{figure*}

Finally, we further examine the effect of access to context inference probabilities compared to most likely context inferences as a function of the disengagement increment parameter $\epsilon_d$ and disengagement decay parameter $\delta_d$. These results are presented in the supplemental material in Figure 2. These results show that context probabilities dominate most likely contexts over a wide range of disengagement dynamics. However, the performance difference tends to be larger in cases that lead to a greater chance of disengagement events occurring. This corresponds to larger values of the disengagement risk increment parameter value $\epsilon_d$ and smaller values of the disengagement risk decay parameter value $\delta_d$. 

\textbf{The Effect of Partial Observability:}
To study the effect of partial observability, we repeat the primary experiments presented in the previous two sections but under the scenario
where the agents do not have access to the $h_t$ and $d_t$ state variables. Instead, the agents are given access to either the most likely context $l_t$ and the time indicator variable $i_t$, or the  context inference probability $\mathbf{p}_t$ and the time indicator variable $i_t$.  We again vary the value of feature uncertainty parameter $\sigma$ from $0$ to $2$ resulting in variation in context inference error from 0\% to approximately 40\%. The results when using the most likely context are given in Figure \ref{fig:perf_reinforce_dqn_ltfn_vs_error}. The results when using context inference probabilities are given in Figure \ref{fig:perf_reinforce_dqn_ptfn_vs_error}.

First, we can see that the performance of the DQN method suffers drastically under partial observability. At a context inference error rate of $0$, the DQN method achieves an average total reward of approximately 1500 under partial observability compared to an average total reward of 3000 with fully observed state. Further, regardless of whether most likely contexts or context probabilities are used, the performance of the DQN agent decays similarly toward an average total reward of approximately 500 at a context inference error rate of approximately 40\%.

We can see a significant contrast when comparing the DQN agent to the REINFORCE agent. The REINFORCE agent experiences a small drop in performance under the $0$ context inference error condition compared to the same condition with fully observed state, thus vastly outperforming the DQN agent. Further, we can see that the REINFORCE agent maintains better performance when using context inference probabilities compared to when using most likely context under partial observability. 

We again perform unpaired t-tests to formally contrast the DQN agent with the REINFORCE agent for each context inference error rate. The performance differences are highly statistically significant with large differences in mean performance across all context inference error rates. These results are presented in Table 1 in the supplemental material.

\vspace{1em}

%-------------------------------------------

\textbf{Sample Complexity of Learning:}
In this experiment, we compare sample learning curves of the DQN and REINFORCE agents for scenarios C-H-D, P-H-D and P-T to illustrate their convergence properties as a function of the number of episodes of training. The results are shown in Figure \ref{fig:train_dqn_vs_reinforce} using a moving average window of $100$ episodes. As expected, REINFORCE exhibits higher variability during learning and takes much longer to converge than the DQN agent. In general, policy gradient methods are known to be less sample efficient than value function methods, which can benefit from off-policy learning using a replay buffer. However, REINFORCE converges at a similar rate and to similar performance in both the P-H-D and P-T scenarios while  the DQN method converges at a similar rate but to much worse performance under the P-T scenario.

\section{Conclusions}\label{sec:conclusions}

In this paper we have investigated the impact of context inference error and partial observability on the ability to learn intervention option selection policies for Just-In-Time adaptive interventions using RL methods. We have introduced a novel simulation environments that captures key aspects of JITAIs including habituation and disengagement risk as well as uncertainty and error in context inferences. We have investigated learning policies which rely on most likely inferred context (as is typically the case in current JITAIs), and have shown that the use of context probabilities significantly outperforms the use of most likely context inferences. We have further shown that there is a stark difference in performance between policy gradient methods and Q-learning methods under partial observability. 

As noted in Section \ref{subsec:rl-jitais} this work has a number of important limitations. First, our primary goal is to quantify the fundamental limits of policy learnability under context inference error and uncertainty as well as partial observability using policy gradient and Q-learning methods. In doing so we have not constrained the RL methods to a realistic number of episodes during learning. As a result, our findings should be interpreted as providing upper bounds on performance in these important and previously unexplored settings. 

Going forward, more work is required to compose the findings of this paper with regard to the use of probabilistic context inference representations with prior work such as \cite{liao2020personalized}, which focuses on sample efficiency of learning. We also note that the drastic loss of performance experienced by traditional Q-learning methods in our experiments may be addressable using state augmentation methods such as the addition of memory or the use of recurrent neural networks that have been proposed in prior work to deal with partial observability. Another potentially interesting possibility is the incorporation of probabilistic dynamic latent variable models to provide beliefs over the full state including psychological latent variables.  

Finally, we note that while the simulation environment was designed to model key issues with context uncertainty and delayed effect of actions, it is limited in other aspects. Nevertheless we believe that the insights we derive have important implications for the development of RL methods that can be applied to improve the effectiveness of real-world JITAIs.

\begin{acknowledgements}
This work was supported by National Institutes of Health Office of Behavior and Social Sciences, National Cancer Institute, and National Institute of Biomedical Imaging and Bioengineering through grants U01CA229445 and 1P41EB028242.  
\end{acknowledgements}

\bibliography{sources}

\appendix
\section{Additional Results}

\subsection{Action Selection Analysis for DQN}

\begin{figure*}[b]
    \centering
    \includegraphics[height=3cm]{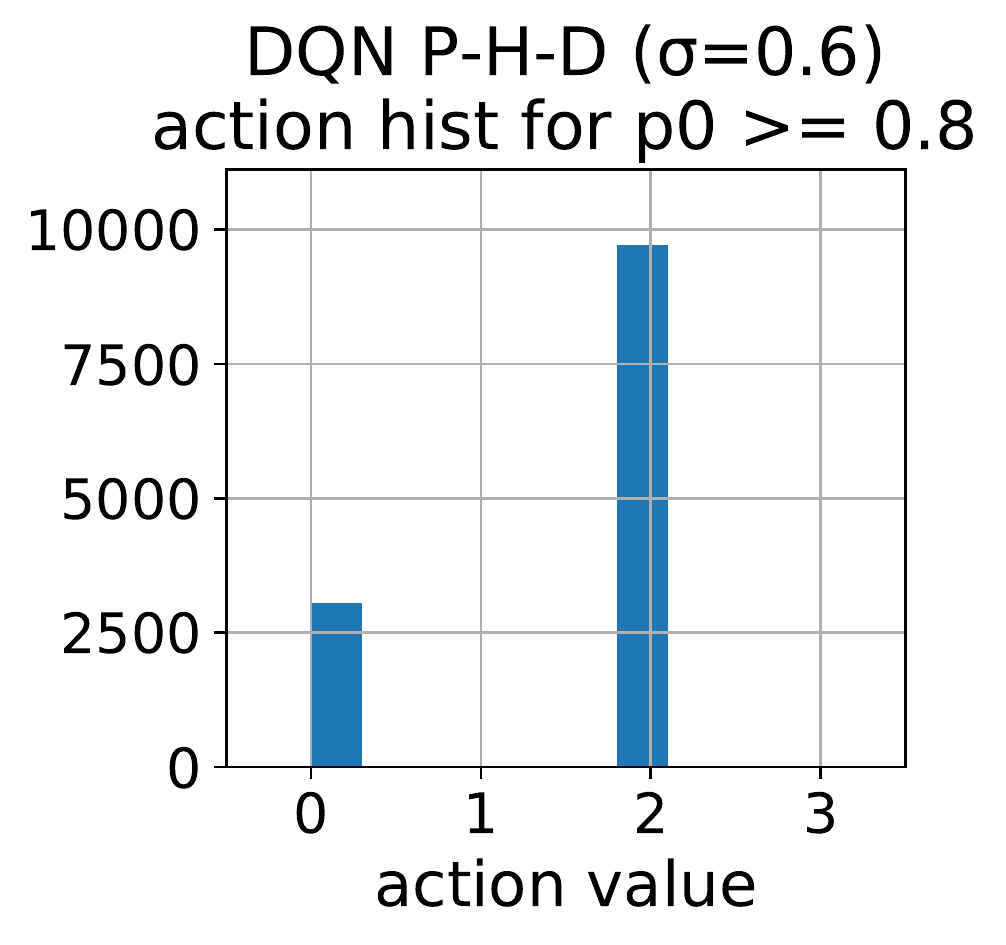}
    \includegraphics[height=3cm]{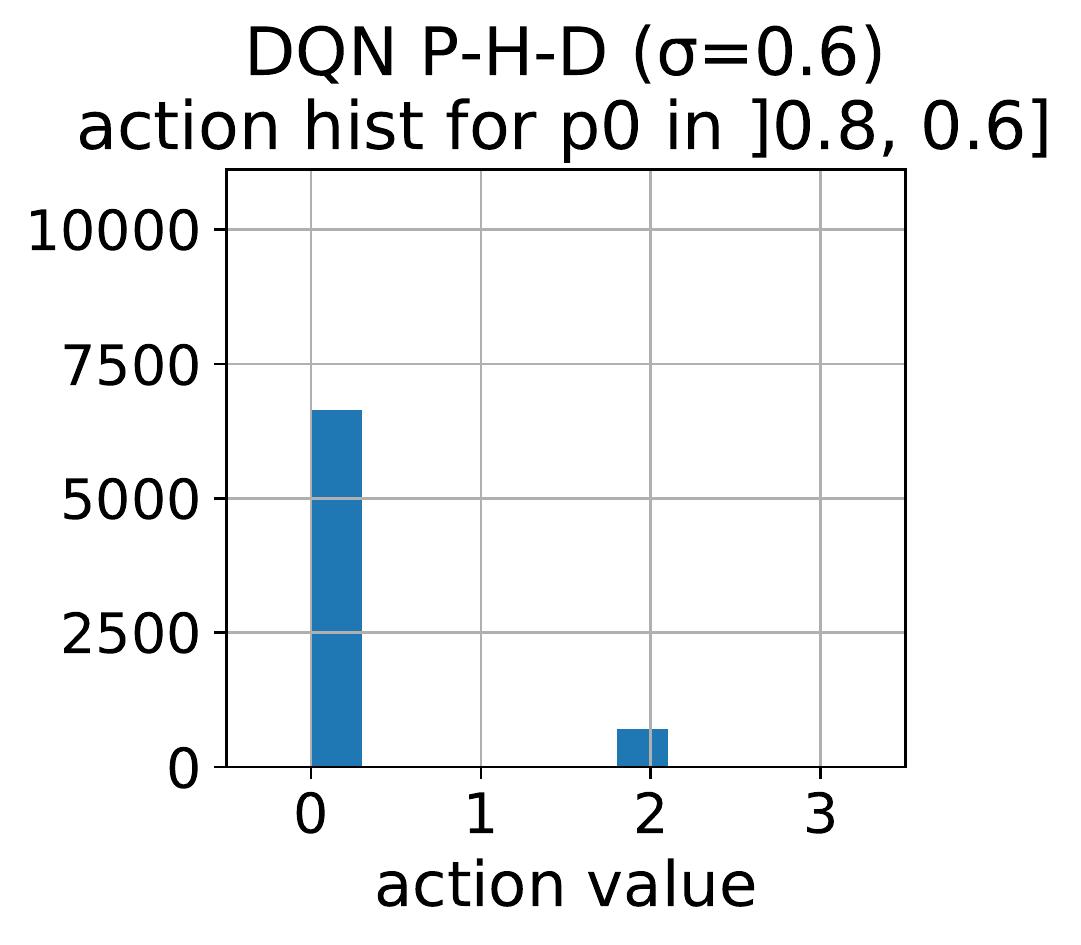}
    \includegraphics[height=3cm]{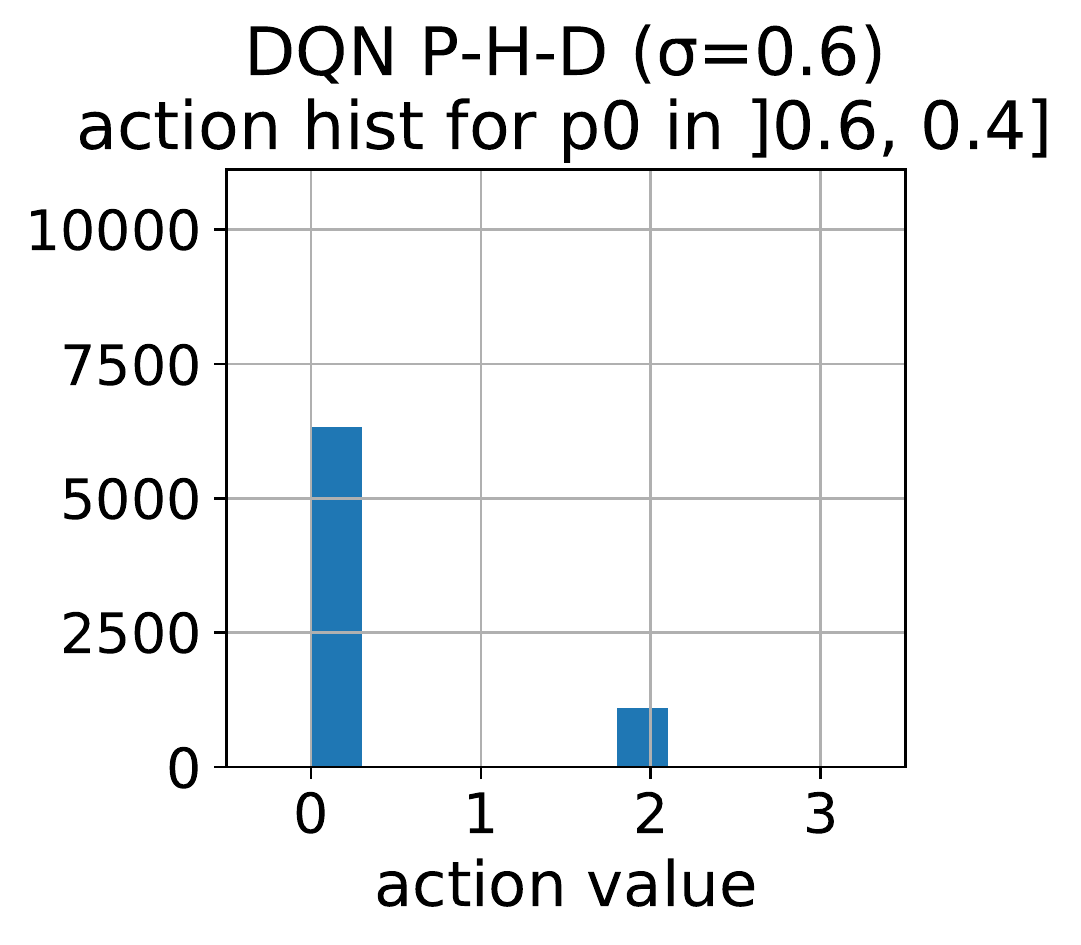}
    \includegraphics[height=3cm]{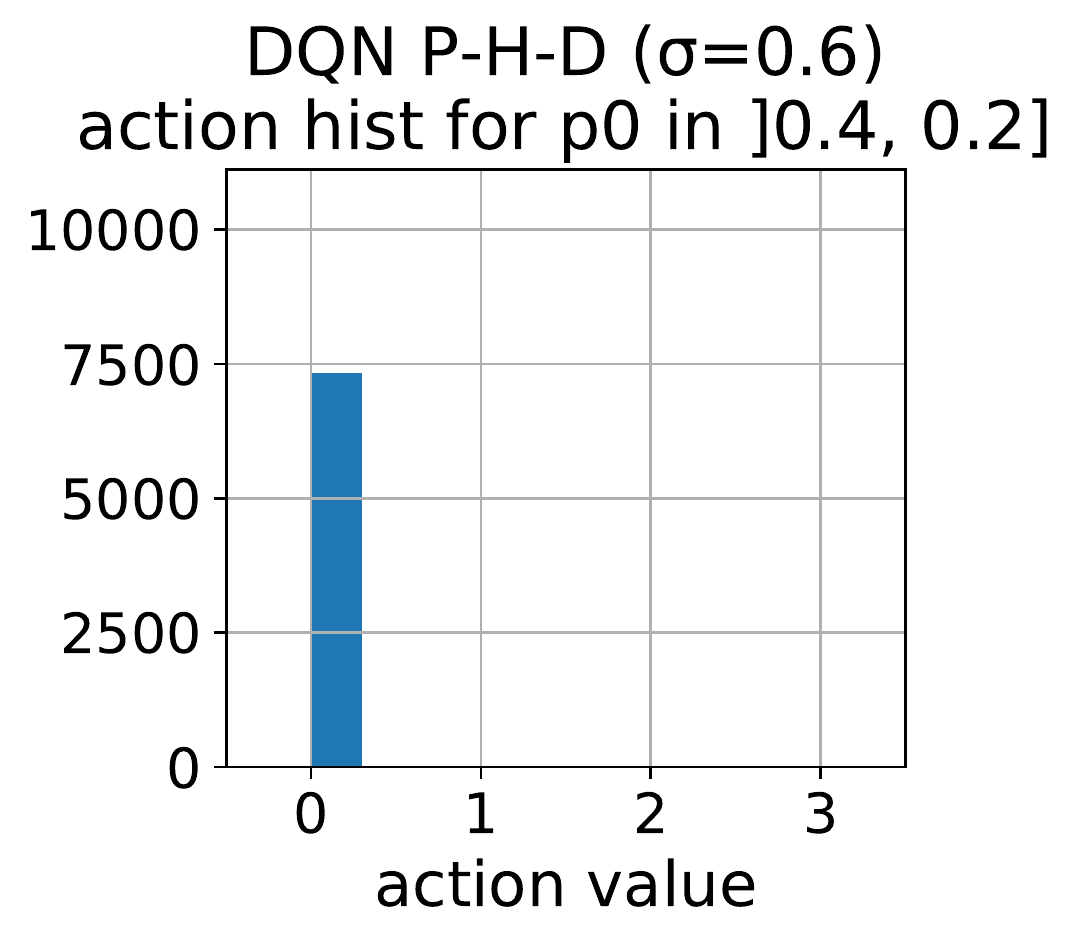}
    \includegraphics[height=3cm]{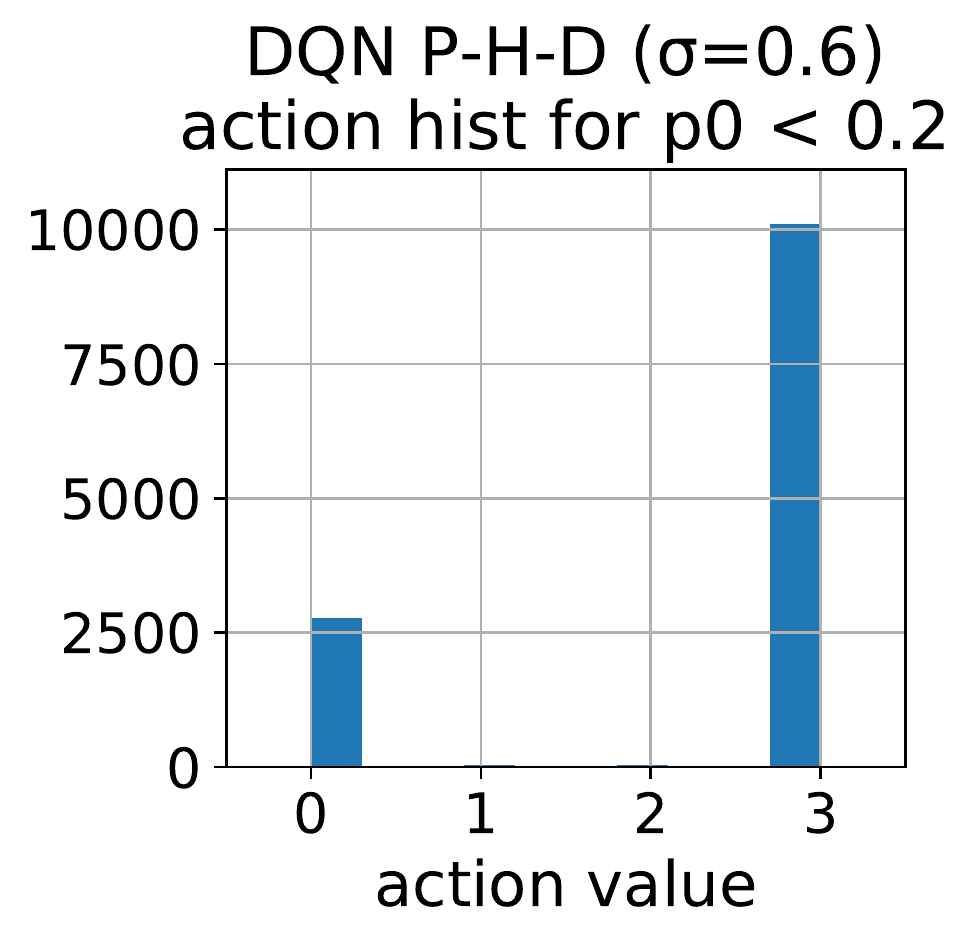}\\
    \includegraphics[height=3cm]{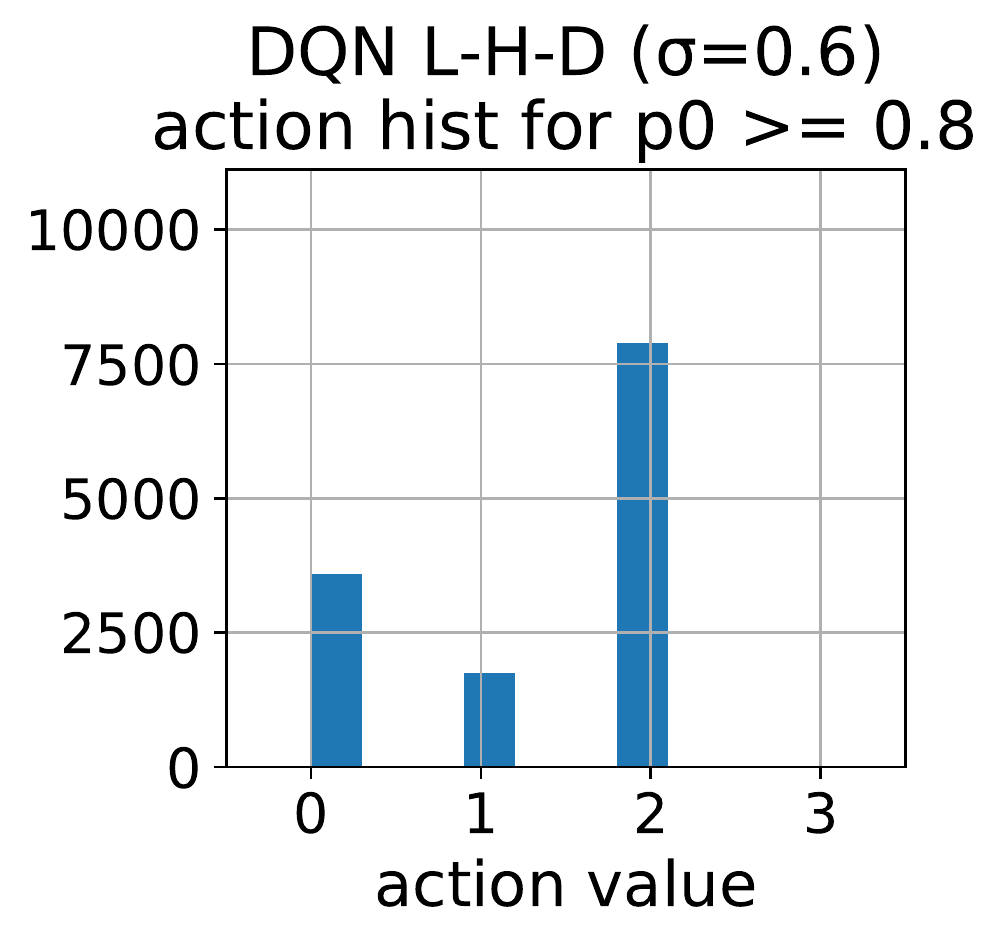}
    \includegraphics[height=3cm]{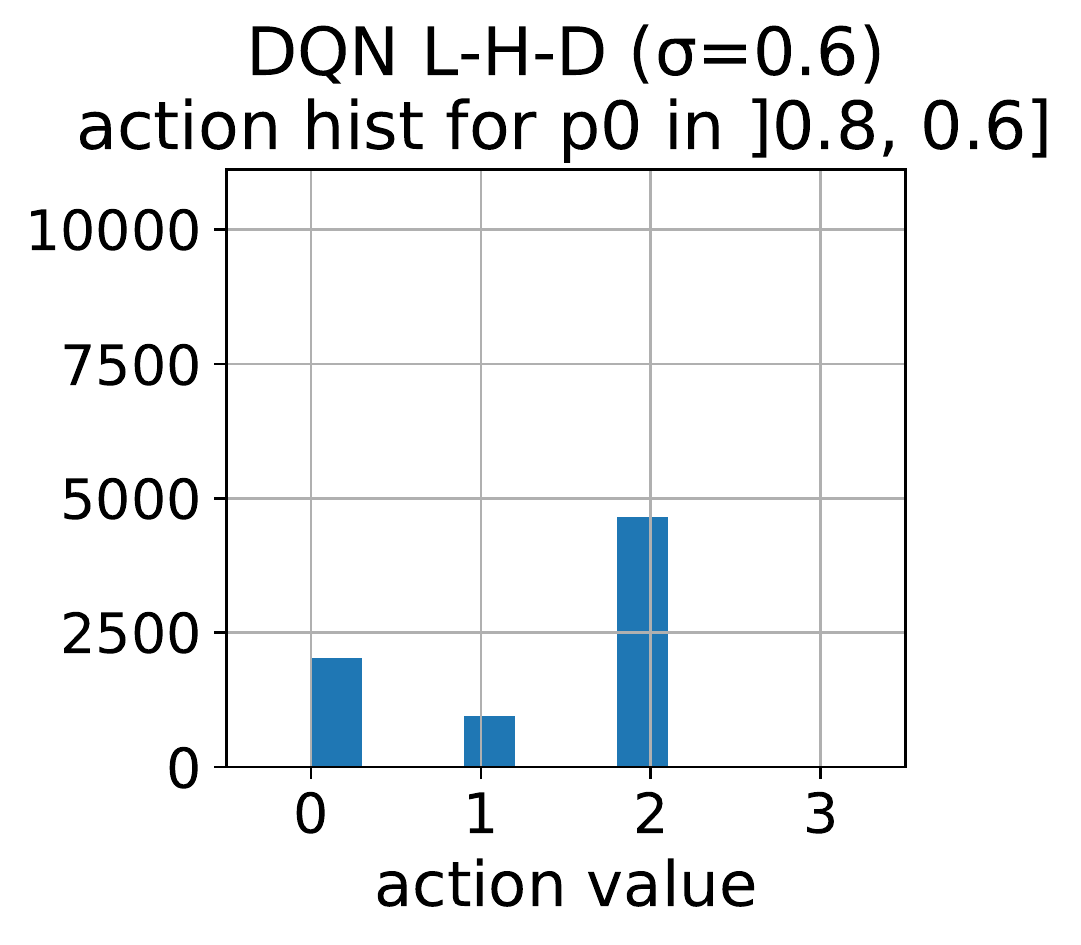}
    \includegraphics[height=3cm]{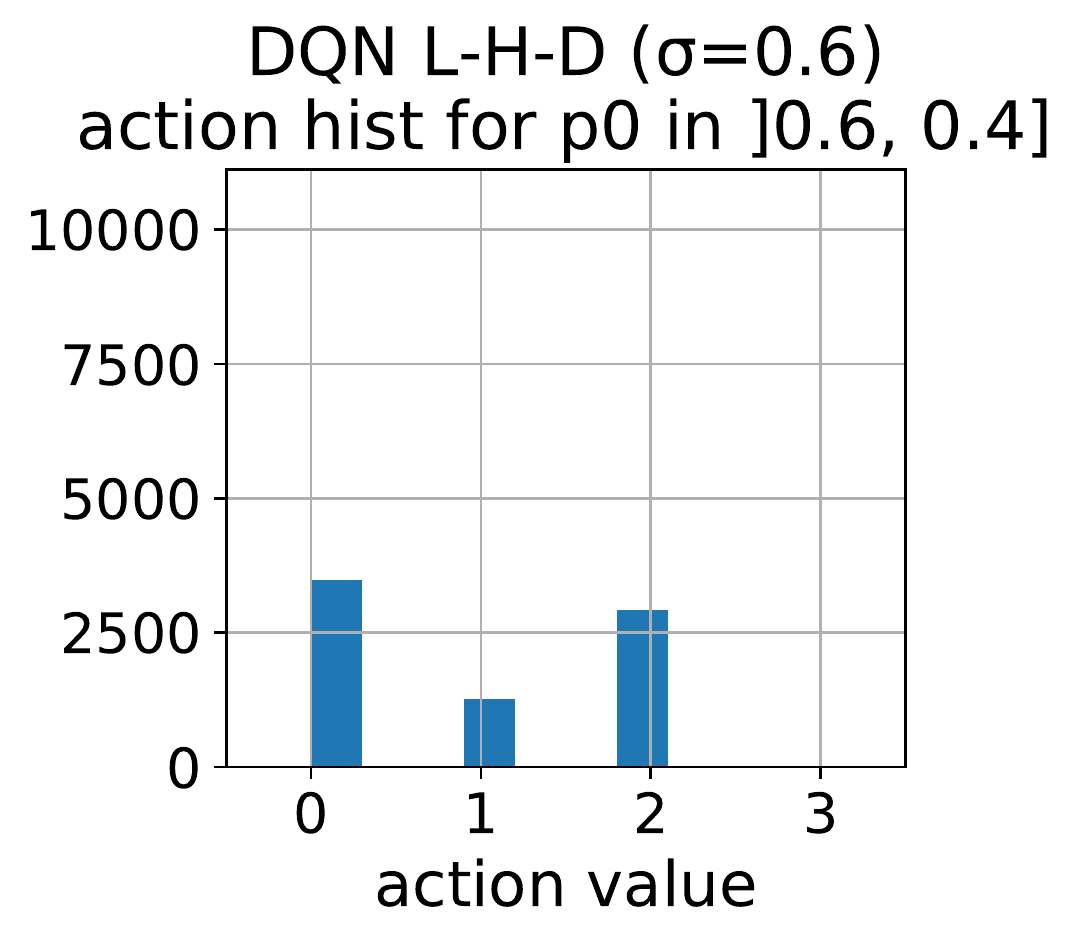}
    \includegraphics[height=3cm]{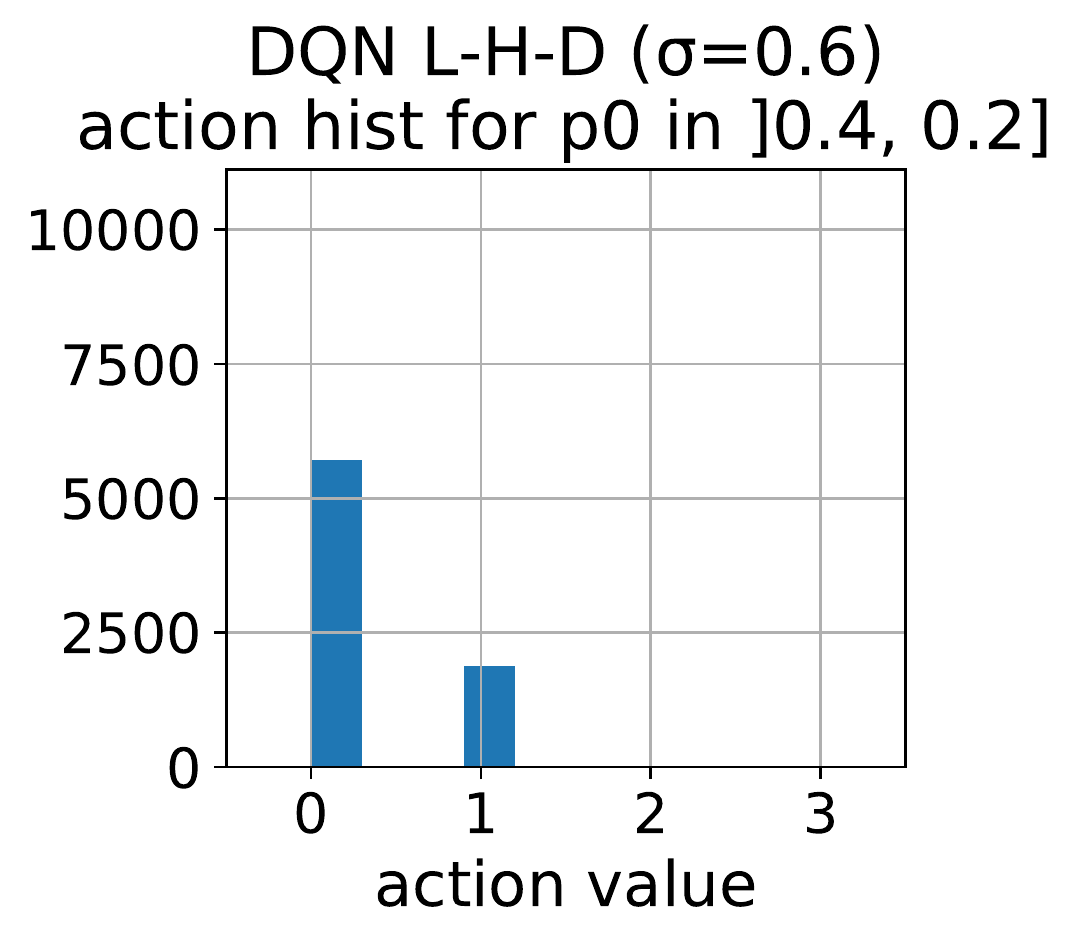}
    \includegraphics[height=3cm]{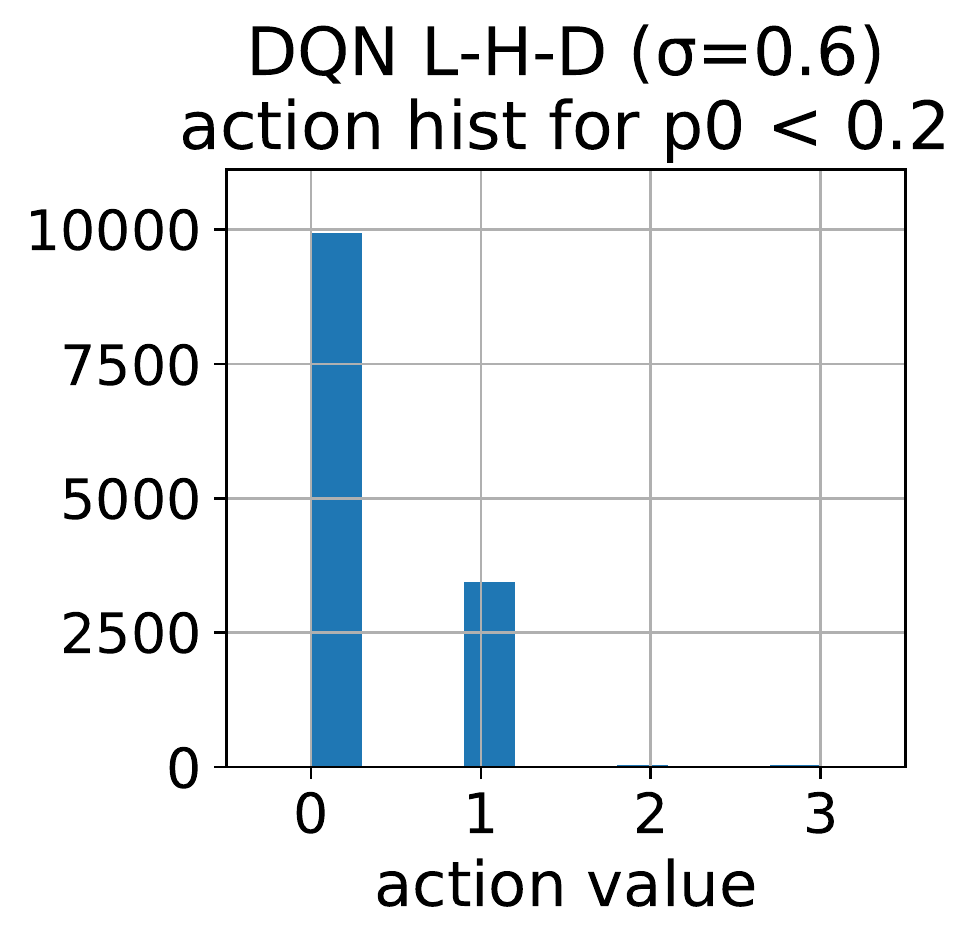}
    \caption{The top row of plots shows
    the distribution of actions selected by DQN when given access to context probabilities. The bottom row of plots shows
    the distribution of actions selected by DQN when given access only to the inferred most likely context. }
        \label{fig:dqn-actions}
\end{figure*}

Figure \ref{fig:dqn-actions} shows the distriubtion of actions taken by the DQN agent.
The top row of plots shows the distribution of actions selected by DQN when given access to context probabilities. The bottom row of plots shows the distribution of actions selected by DQN when given access only to the inferred most likely context. Each plot in each row corresponds to the distribution of actions in a specific range of context inference probabilities. All results are for a context inference error rate of 17\%.

%----------------------------------------------------

\subsection{Statistical Significance of Performance Differences Under Partial Observability}

We perform unpaired t-tests to formally contrast the DQN agent with the REINFORCE agent for each context inference error rate under the partial observability condition. The performance differences are highly statistically significant with large differences in mean performance across all context inference error rates. These results are presented in Table \ref{tab:unpaired t-tests2}.

\begin{table}[h]
  \centering
  \caption{Unpaired t-tests on performance for scenarios REINFORCE L-T vs. DQN L-T, and scenarios REINFORCE P-T vs. DQN P-T, for different inferred error rates. Effect is the difference of the average returns.}
  \label{tab:unpaired t-tests2}
  \begin{tabular}{cccc}
    \toprule
    \bfseries REINF. vs. DQN & \bfseries Error Rate & \bfseries Effect & \bfseries p-value \\
    \midrule
    L-T & 10\%  & 1056.15 & 0.00002 \\
    L-T & 17\%  & 745.50  & 0.00009 \\
    L-T & 27\%  & 450.42  & 0.00005 \\
    L-T & 31\%  & 394.75  & 0.00139 \\
    L-T & 41\%  & 445.21  & 0.01435 \\
    \midrule
    P-T & 10\%  & 1367.03  & 0.00002 \\
    P-T & 17\%  & 1032.63  & 0.00347 \\
    P-T & 27\%  & 927.28   & 0.00008 \\
    P-T & 31\%  & 681.53   & 0.00054 \\
    P-T & 41\%  & 329.25   & 0.00014 \\
    \bottomrule
  \end{tabular}
\end{table}

%----------------------------------------------------

\subsection{Performance as a Function of Disengagement Dynamics Parameters.}

\begin{figure*}[ht]
    \centering
    \includegraphics[height=5cm]{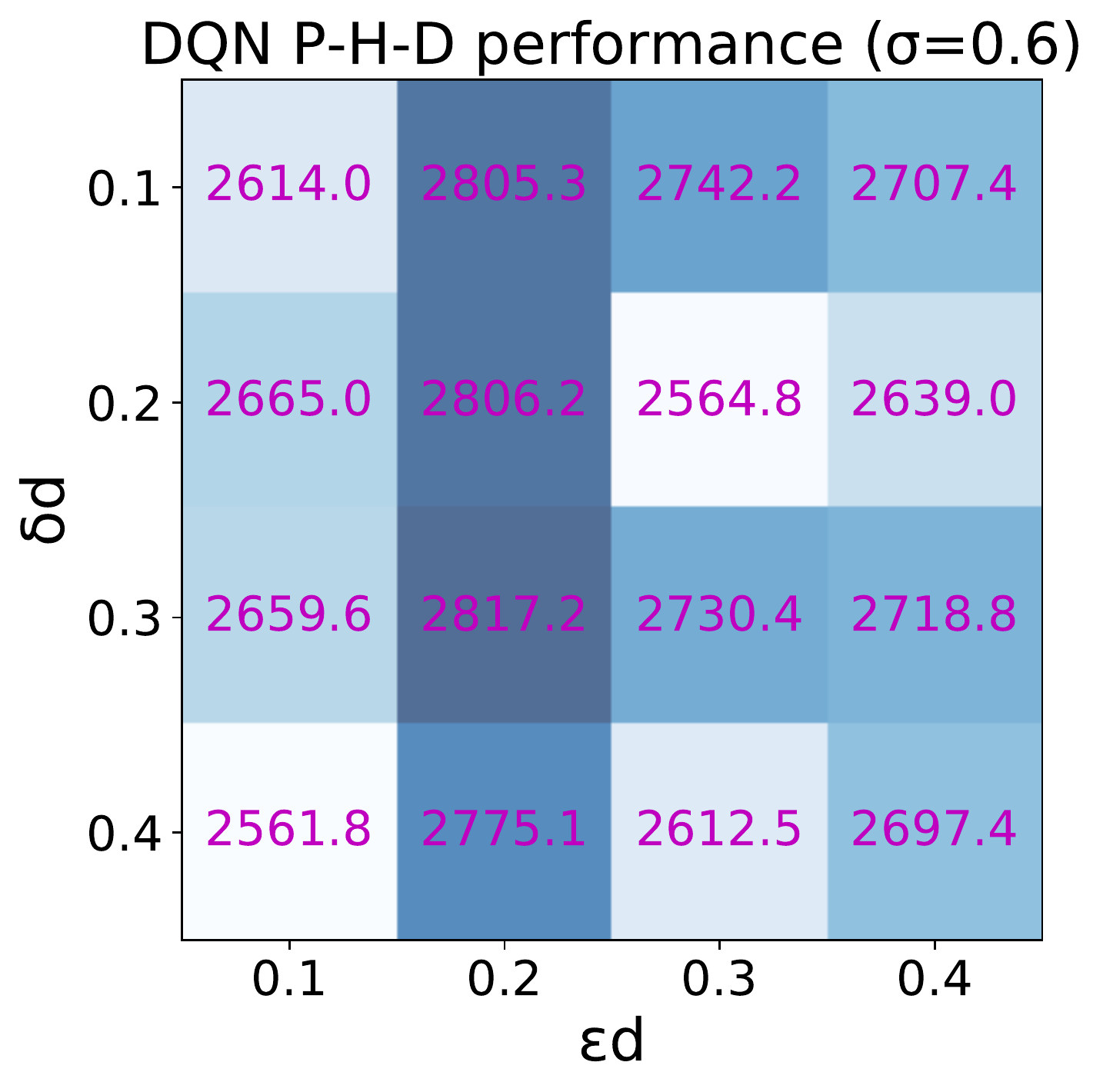}
    \hspace{.5cm}
    \includegraphics[height=5cm]{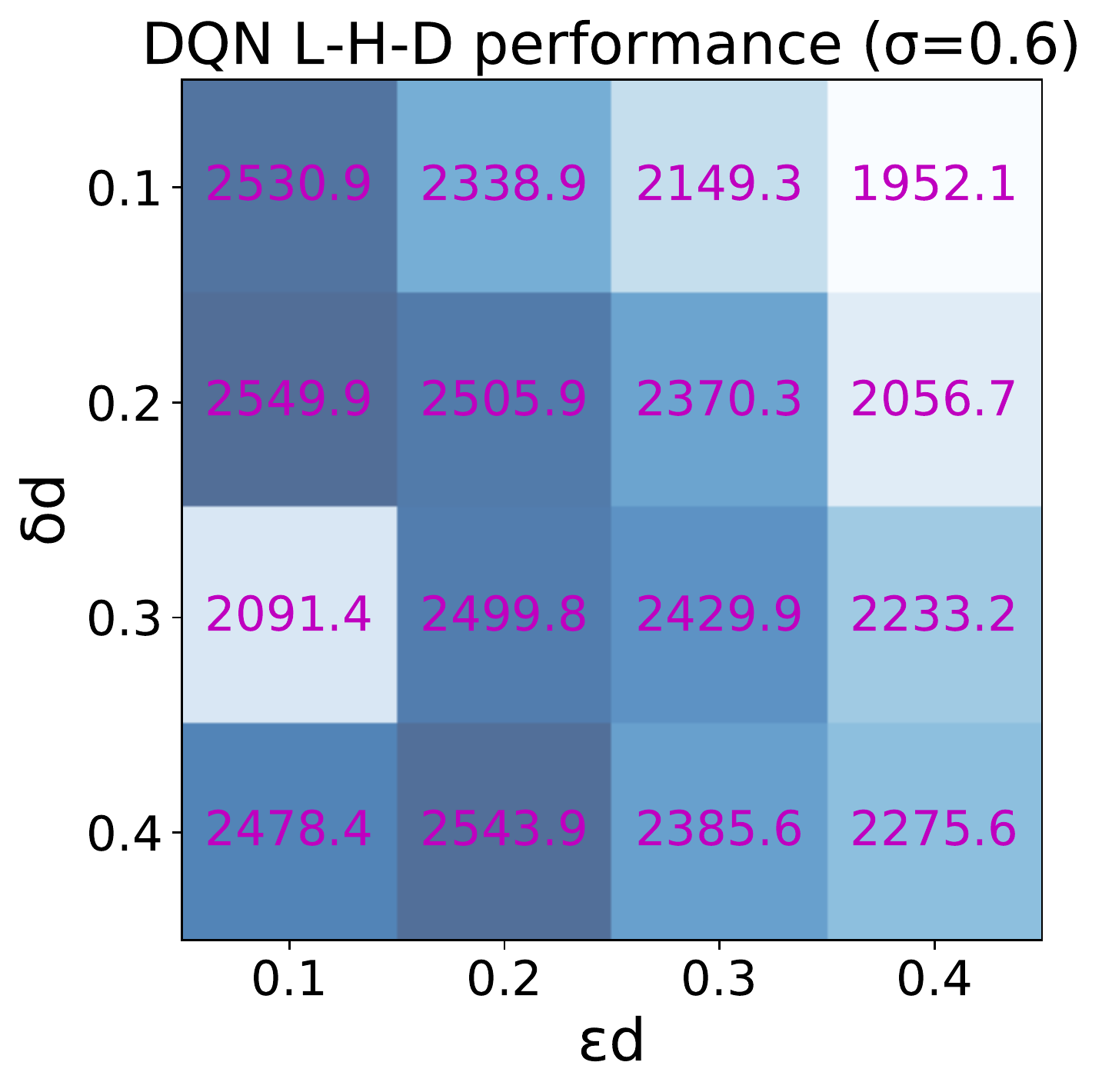}
    \hspace{.5cm}
    \includegraphics[height=5.35cm]{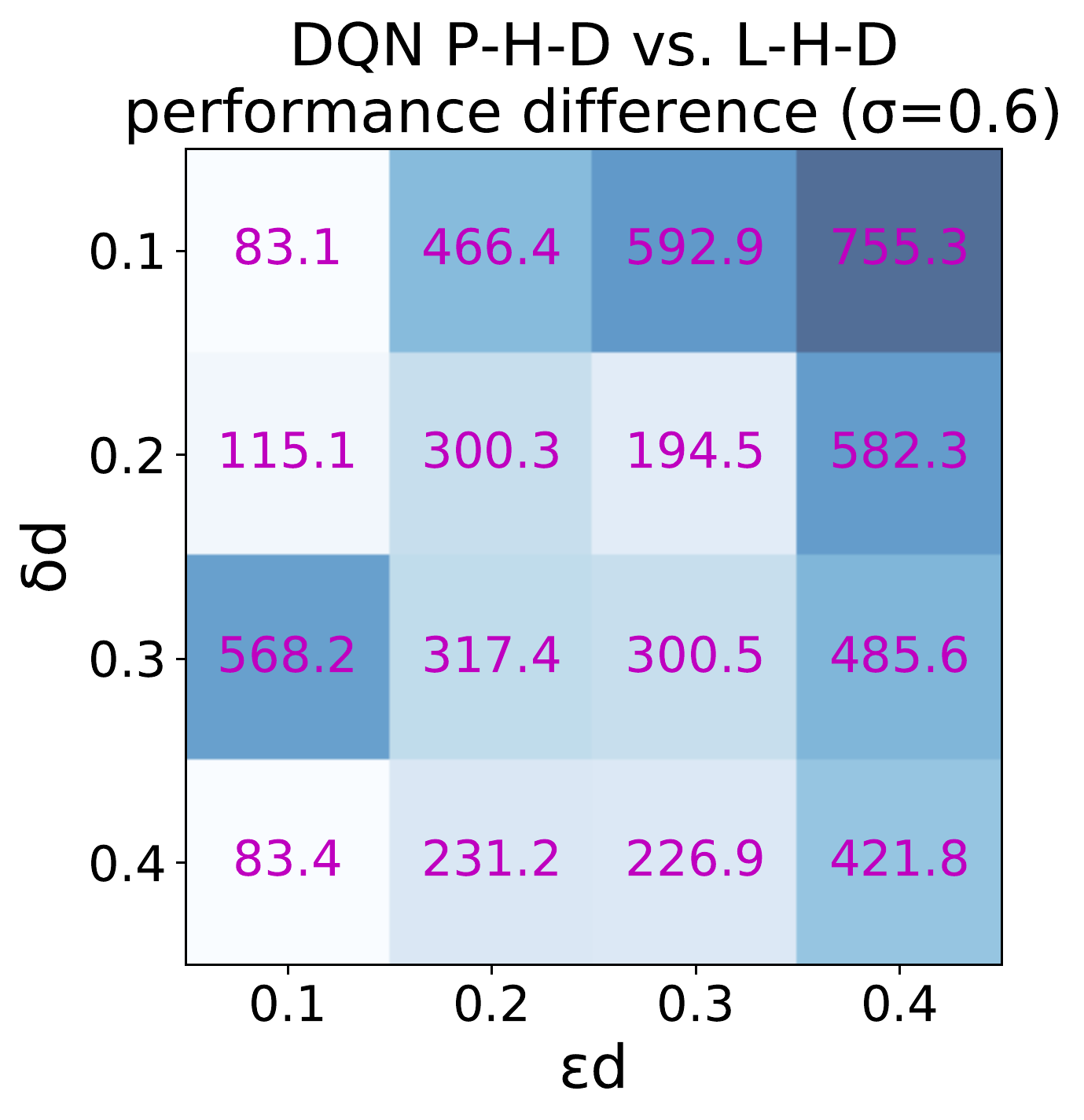}
    \includegraphics[height=5cm]{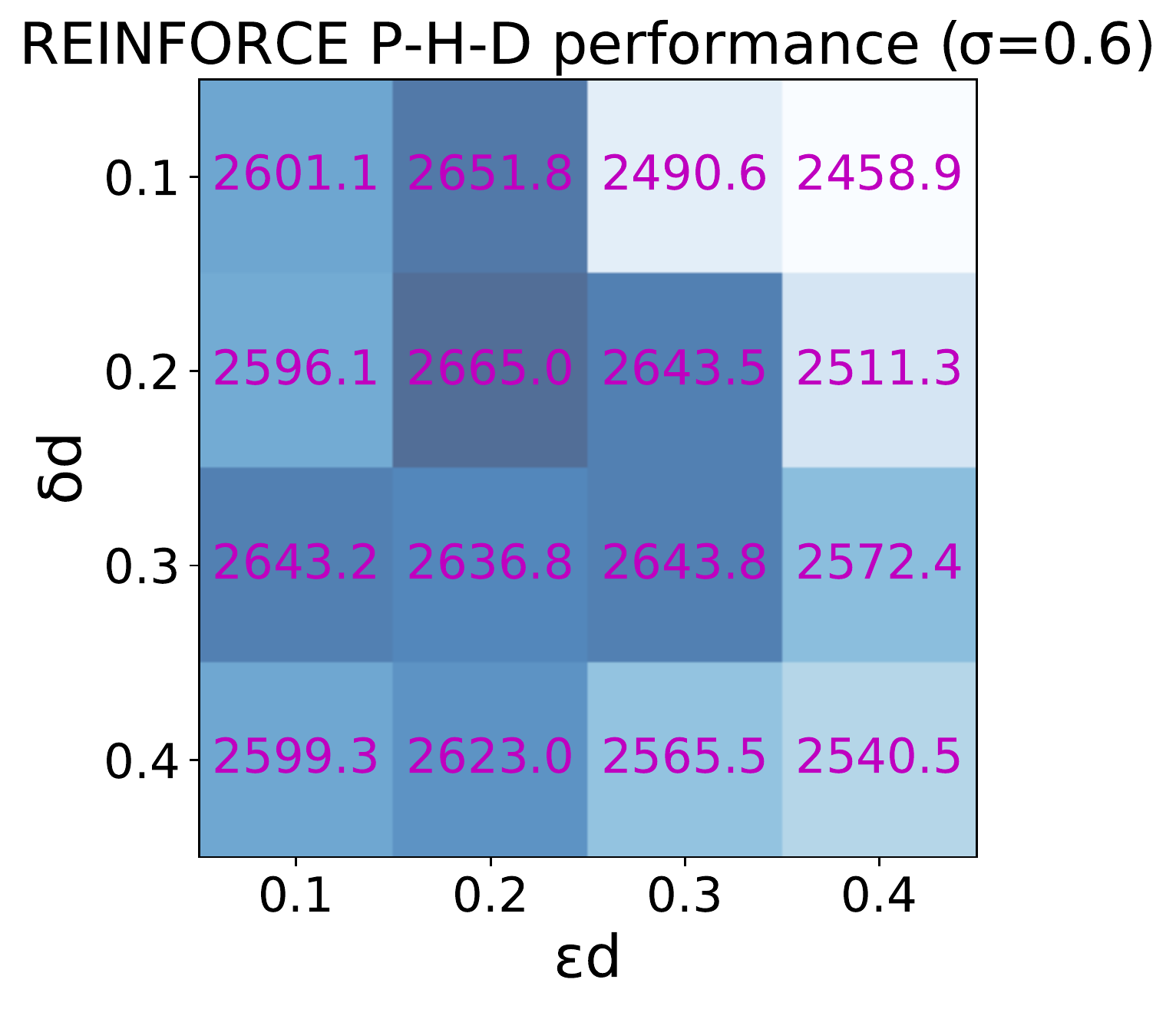}
    \includegraphics[height=5cm]{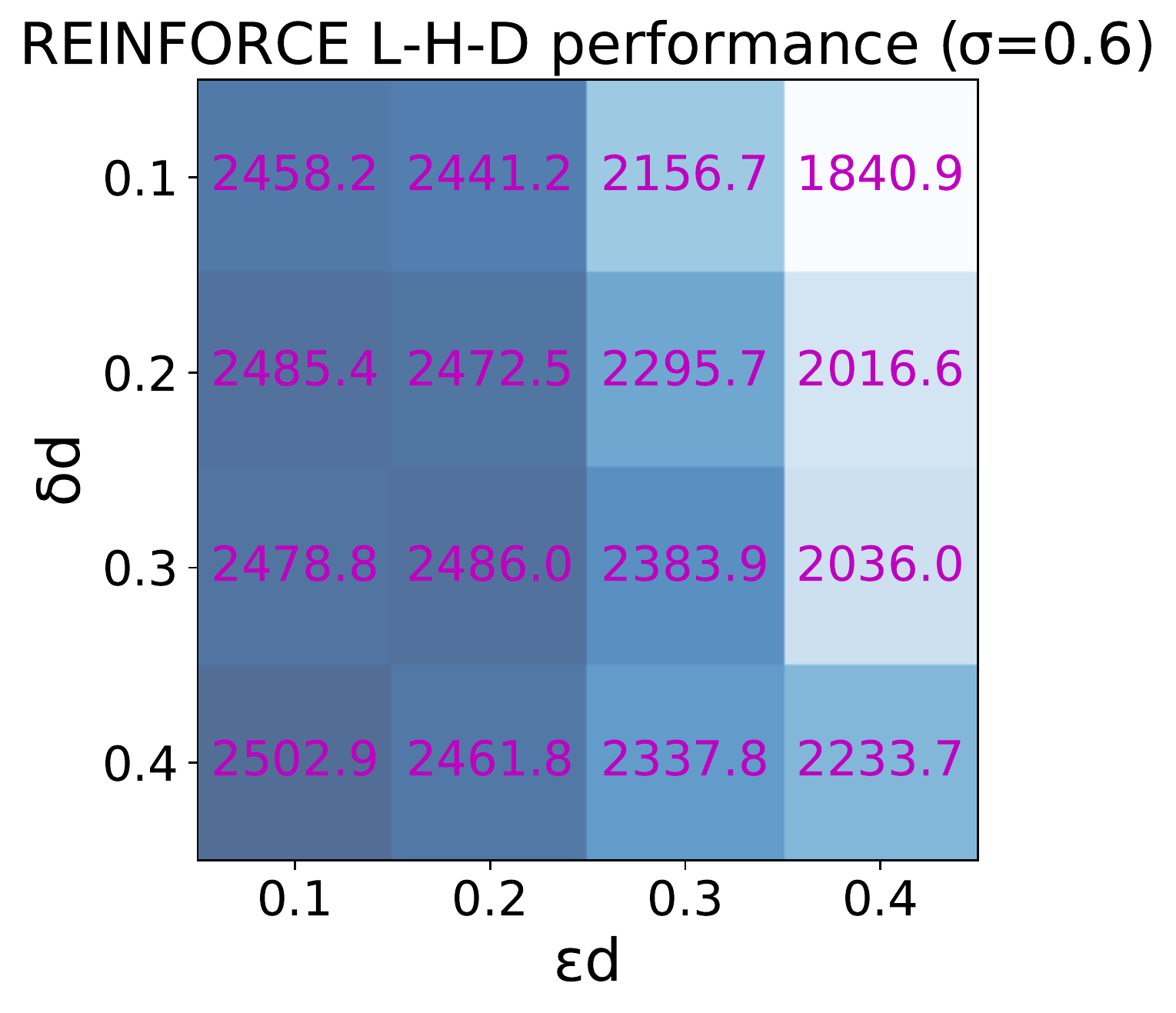}
    \includegraphics[height=5.35cm]{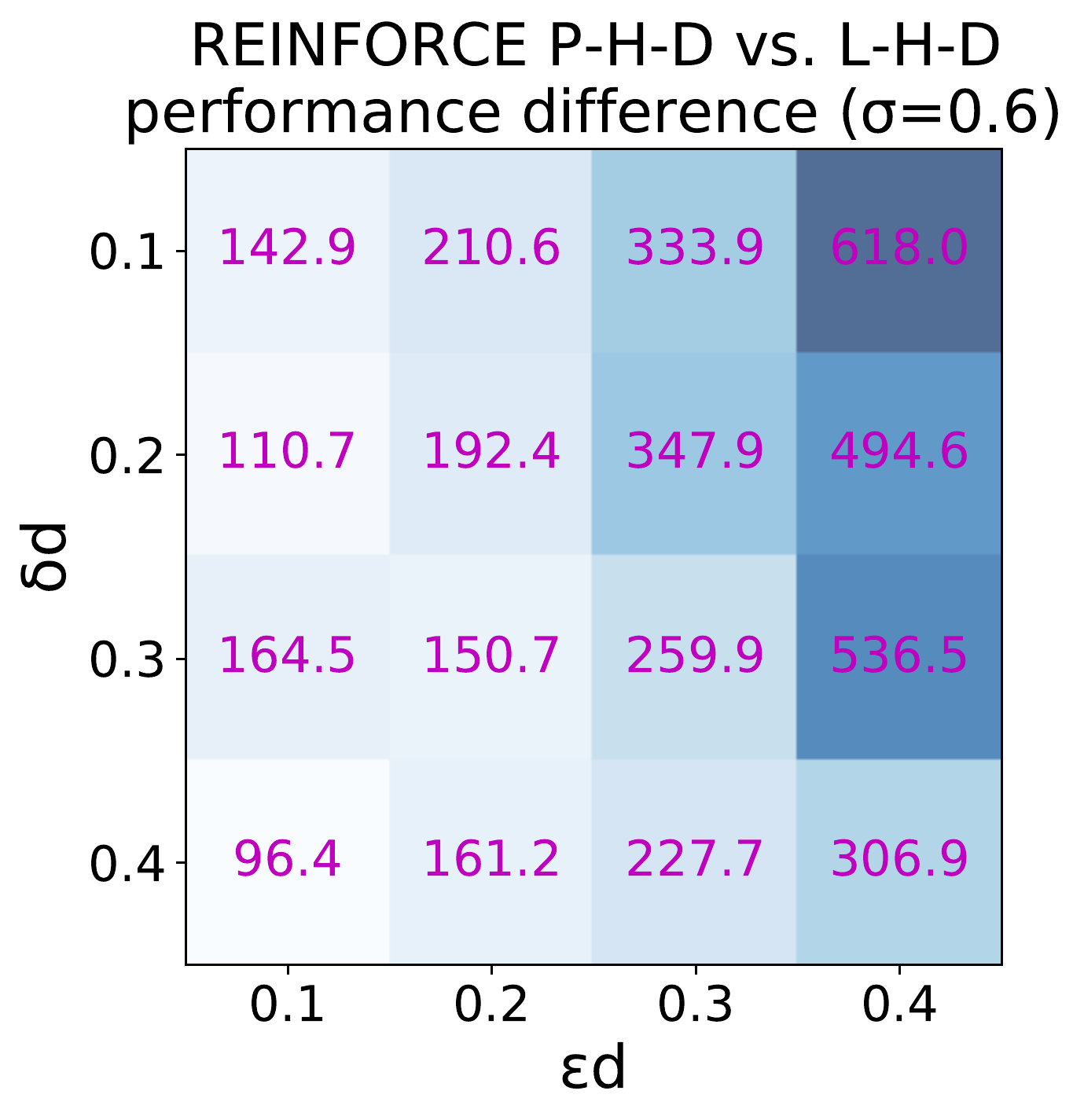}
    \caption{Performance as a function of the disengagement increment $\epsilon_d$ and decay parameters $\delta_d$, for DQN (top row) and REINFORCE (bottom row).}
    \label{fig:DQN and REINFORCE heatmap}
\end{figure*}

For both agents, we study how the performance of learned policies varies as a function of the disengagement increment parameter $\epsilon_d$ and disengagement decay parameter $\delta_d$. The presented results correspond to $\sigma = 0.6$ and habituation and disengagement observed. The results are given in Figure \ref{fig:DQN and REINFORCE heatmap}. As we can see, these results show that the use of context inference probabilities improves on using most likely context inference over a wide range of settings of these variables. However, the performance difference tends to be larger in cases that lead to a greater chance of disengagement events occurring. This corresponds to larger values of the disengagement risk increment  parameter value $\epsilon_d$ and smaller values of the disengagement risk decay parameter value $\delta_d$.

\end{document}